%% file: sample-sigconf.tex
\documentclass[sigconf]{acmart}

\usepackage{graphicx}
\usepackage{appendix}
\usepackage{tikz}
\usepackage{comment}
\usepackage{amsmath}
\usepackage{color}
\usepackage{colortbl}
\usepackage{bbm}
\usepackage{makecell}
\usepackage{pifont}
\usepackage{multirow}
\usepackage{wrapfig}
\usepackage{hyperref}
\usepackage{balance}

\definecolor{LightGray}{gray}{0.9}
\newcommand{\cmark}{\ding{51}}
\newcommand{\xmark}{\ding{55}}
\newcommand{\JFm}{$\mathcal{J} \& \mathcal{F}_{m}$}
\newcommand{\Jm}{$\mathcal{J}_{m}$}
\newcommand{\Jr}{$\mathcal{J}_{r}$}
\newcommand{\Fm}{$\mathcal{F}_{m}$}
\newcommand{\Fr}{$\mathcal{F}_{r}$}

\AtBeginDocument{%
  \providecommand\BibTeX{{%
    \normalfont B\kern-0.5em{\scshape i\kern-0.25em b}\kern-0.8em\TeX}}}

\copyrightyear{2022}
\acmYear{2022}
\setcopyright{acmlicensed}
\acmConference[MM '22] {Proceedings of the 30th ACM International Conference on Multimedia }{October 10--14, 2022}{Lisboa, Portugal.}
\acmBooktitle{Proceedings of the 30th ACM International Conference on Multimedia (MM '22), October 10--14, 2022, Lisboa, Portugal}
\acmPrice{15.00}
\acmISBN{978-1-4503-9203-7/22/10}
\acmDOI{https://doi.org/10.1145/3503161.3547909}

\acmSubmissionID{666}

\begin{document}

\title{In-N-Out Generative Learning for Dense Unsupervised Video Segmentation}









\author{Xiao Pan$^{1,2}$, Peike Li$^{2,3}$, Zongxin Yang$^{1}$, Huiling Zhou$^{2}$, Chang Zhou$^{2}$,\\ Hongxia Yang$^{2}$, Jingren Zhou$^{2}$, Yi Yang$^{1}$
} 
\affiliation{%
$^{1}$ReLER Lab, CCAI, Zhejiang University \country{China} \\
$^{2}$Alibaba DAMO Academy \country{China} \\
$^{3}$Australian Artificial Intelligence Institute, University of Technology Sydney \country{Australia} \\
}

\renewcommand{\shortauthors}{Xiao Pan et al.}

\begin{abstract}
\input{section/0_abstract}
\end{abstract}

\ccsdesc[500]{Computing methodologies~Video segmentation}
\ccsdesc[500]{Computing methodologies~Image segmentation}
\ccsdesc[500]{Computing methodologies~Image representations}


\keywords{unsupervised video object segmentation, self-supervised learning, dense prediction, generative learning}


\maketitle

\section{Introduction}

\input{section/1_introduction}

\section{Related Work}
\input{section/2_related_work}

\section{Method}
\input{section/3_method}

\section{Experiment}

\input{section/4_experiment}

\section{Conclusion}
\input{section/5_conclusion}

\begin{acks}
Xiao Pan and Zongxin Yang were in part supported by the Fundamental Research Funds for the Central Universities (No. 226-2022-00087).
\end{acks}

\clearpage
\bibliographystyle{ACM-Reference-Format}
\balance 
\bibliography{acmart}

\clearpage
\appendix
\input{section/appendix}

\end{document}

%% file: section/0_abstract.tex

In this paper, we focus on unsupervised learning for Video Object Segmentation (VOS) which learns visual correspondence (\textit{i.e.}, the similarity between pixel-level features) from unlabeled videos.
Previous methods are mainly based on the contrastive learning paradigm, which optimize either in image level or pixel level. Image-level optimization (\textit{e.g.}, the spatially pooled feature of ResNet) learns robust high-level semantics but is sub-optimal since the pixel-level features are optimized implicitly. By contrast, pixel-level optimization is more explicit, however, it is sensitive to the visual quality of training data and is not robust to object deformation. 
To complementarily perform these two levels of optimization in a unified framework, we propose the In-aNd-Out (INO) generative learning from a purely generative perspective with the help of naturally designed class tokens and patch tokens in Vision Transformer (ViT).
Specifically, for image-level optimization, we force the out-view imagination from local to global views on class tokens, which helps capture high-level semantics, and we name it as out-generative learning. As to pixel-level optimization, we perform in-view masked image modeling on patch tokens, which recovers the corrupted parts of an image via inferring its fine-grained structure, and we term it as in-generative learning. 
To discover the temporal information better, we additionally force the inter-frame consistency from both feature and affinity matrix levels.
Extensive experiments on DAVIS-2017 \texttt{val} and YouTube-VOS 2018 \texttt{val} show that our INO outperforms previous state-of-the-art methods by significant margins. Code:
\textcolor{red}{\url{https://github.com/pansanity666/INO_VOS}}

%% file: section/1_introduction.tex
\begin{figure}[t]
\small
\centering
\includegraphics[width=0.9\linewidth]{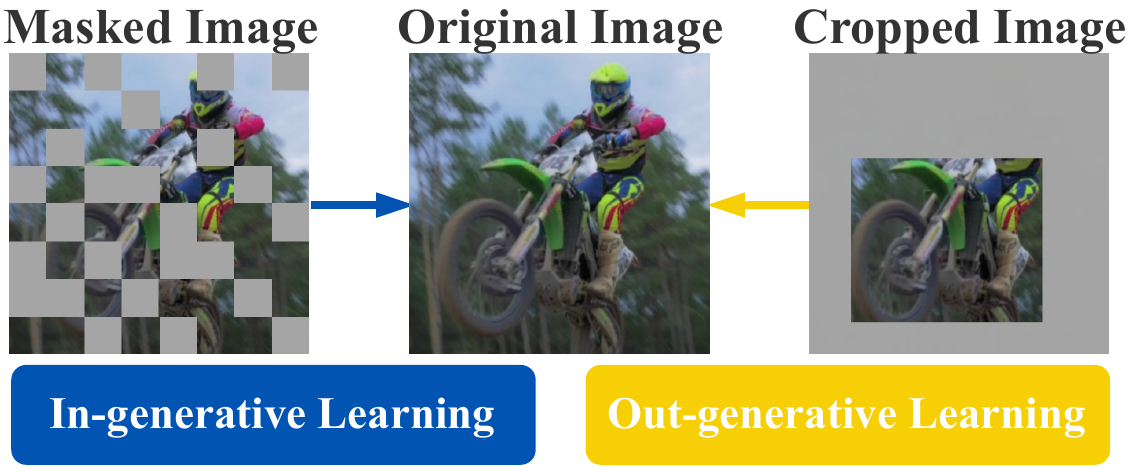}
\vspace{-3mm}
\caption{Idea illustration for In-N-Out generative learning. In-generative learning is the in-view recovery of masked patches, while out-generative learning is the out-view imagination from local to global views.}
\label{fig:idea-illustration}
\vspace{-4mm}
\end{figure}

Video Object Segmentation~\cite{R4_added5_wang2021survey, Dense_araslanov2021dense,VOS1_miao2020memory,STM_oh2019video, CFBI_eccv_yang2020collaborative,R2_added0_yang2019unsupervised} (VOS) is a fundamental video understanding task having a wide range of real-world applications, \textit{e.g.}, augmented reality~\cite{VR_xiong2021augmented}, and video editing~\cite{Video_editing_lin2021automated}.
In this task, we aim to segment a specified object instance throughout the entire video sequence, given only the ground-truth object mask on the first frame.
Although the development of convolutional neural networks (CNNs)~\cite{ResNet_he2016deep} has significantly advanced the VOS task, the success of these approaches highly relies on the cost-intensive and time-consuming dense mask annotations to train the networks.
Moreover, the fully-supervised VOS is also largely limited by the diversity and scale of the annotated datasets.
To relieve such limitations, recently unsupervised/self-supervised methods for VOS~\cite{Dense_araslanov2021dense,VRW_jabri2020space,MAST_lai2020mast} have drawn considerable attention, which can completely liberate the request for mask annotations.

However, the previous approaches still suffer from the following perspectives:
(i)\textit{Image-level vs. pixel-level}. Several existing methods~\cite{VRW_jabri2020space,Rethinking_xu2021rethinking} perform self-supervised learning on the image-level features, \textit{e.g.}, the spatially pooled feature of ResNet~\cite{ResNet_he2016deep}. Such optimization can learn robust high-level semantics, however, it is sub-optimal since the pixel-level features used for calculating the final correspondence are learned implicitly. 
Another line of works \cite{Dense_araslanov2021dense,MAST_lai2020mast,TimeCycle_wang2019learning} explicitly optimize on the pixel-level features, however, such paradigm may be sensitive to the visual quality of training data and is lack of the high-level semantic scope which may reduce the robustness toward deformation. We believe that these two levels of optimization are complementary to each other and can be integrated into a unified framework; 
(ii) \textit{CNN vs. ViT}. Previous unsupervised methods~\cite{Dense_araslanov2021dense,VRW_jabri2020space,Rethinking_xu2021rethinking} for VOS adopt the convolutional networks (CNNs)~\cite{ResNet_he2016deep} as the backbone model. However, we propose that  Vision Transformer (ViT)~\cite{ViT_dosovitskiy2020image,DeiT_touvron2021training} is a better choice for unifying these two levels of optimization. Specifically, different from the CNNs which obtain the image-level features via the average pooling of pixel-level features, the naturally designed class tokens and patch tokens of ViT have their own semantic meanings, \textit{i.e.}, class tokens capture the high-level semantics which are suitable for image-level optimization, while patch tokens represent fine-grained details, which are appropriate for pixel-level optimization. 
(iii) \textit{Contrastive vs. generative}. Previous self-supervised methods for VOS~\cite{Dense_araslanov2021dense,VRW_jabri2020space,Rethinking_xu2021rethinking} employ a self-training objective mainly based on contrastive formulation~\cite{MOCO_he2020momentum}. However, their experimental results show unsatisfactory scalability toward the training data scales, especially for the pixel-level methods~\cite{Dense_araslanov2021dense}. 
On the other hand, the recent success of the generative learning paradigm~\cite{MAE_he2021masked} shows its better scalability and robustness. Therefore, we propose that solving the VOS task from the generative learning perspective is promising, yet unexplored.

To tackle the aforementioned issues, we present a simple yet effective framework called In-aNd-Out (INO) generative learning  from a novel fully generative learning perspective.
As opposed to the previous methods~\cite{Dense_araslanov2021dense,VRW_jabri2020space,Rethinking_xu2021rethinking} (detailed in Table \ref{table:method-comparision}), INO integrates image-level and pixel-level optimization in a unified  framework with the help of naturally designed high-level class tokens and fine-grained patch tokens in ViT.

More concretely, as illustrated in Fig.~\ref{fig:idea-illustration}, our INO framework contains two generative objectives, \textit{i.e.}, in-generative learning and out-generative learning. 
(i) \textit{Out-generative learning} is the \textit{out-view} imagination from local views to global views  on class tokens, which corresponds to the image-level optimization. 
We aim to learn high-level visual semantics via imagining the global visual information given only a portion of the local visual part, which may improve the robustness toward deformation and occlusion.
(ii) \textit{In-generative learning} is the \textit{in-view} recovery of randomly masked patch tokens in the feature embedding space, which belongs to the pixel-level optimization. 
The goal of in-generative learning is to capture fine-grained structural information, which is beneficial for recognizing the gradually appearing parts of a semantic object given incomplete labels in the first frame.
In this way, both in-and-out generative objectives complement each other towards better visual representation learning.
To better discover the temporal information, we additionally equip INO with temporally-persistent constraints, by forcing the inter-frame consistency from both feature level and affinity matrix level. Extensive experiments on DAVIS-2017 \texttt{val} and YouTube-VOS 2018 \texttt{val} show that our INO outperforms previous methods by significant margins.
Our main contributions are summarized as follows,
\vspace{-1mm}
\begin{itemize}
\item We propose a simple yet effective framework INO to tackle the challenging unsupervised learning for VOS, which integrates image-level and pixel-level optimization in a unified  framework by leveraging the structural superiority of ViT.
\item To the best of our knowledge, we make the first attempt to conduct unsupervised learning for VOS from a novel fully generative learning perspective  based on the idea of masked image modeling.
\item We attain a new state-of-the-art performance on unsupervised learning for VOS. 
\end{itemize}
\vspace{-2mm}

\input{table/method-comparision}

%% file: table/method-comparision.tex
\begin{table}[t]
\footnotesize
\caption{Comparisons between different state-of-the-art VOS methods.}
\vspace{-2mm}
\begin{center}
{
\setlength{\tabcolsep}{1.1mm}{
\begin{tabular}{l|c|c|c|c}
\rowcolor{LightGray}
\hline
Method &\makecell{Self-\\supervised} & \makecell{Transformer- \\ based} & \makecell{Unified \\optimization} &  \makecell{Generative \\learning} \\ \hline\hline
AOT (NIPS 2021)~\cite{yang2021associating} & \xmark & \cmark & \xmark & \xmark  \\ \hline
CRW  (NIPS 2020)~\cite{VRW_jabri2020space} & \cmark & \xmark & \xmark &  \xmark \\ \hline
VFS  (ICCV 2021)~\cite{Rethinking_xu2021rethinking} & \cmark & \xmark & \xmark & \xmark \\ \hline
DUL  (NIPS 2021)~\cite{Dense_araslanov2021dense} & \cmark & \xmark & \xmark &  \xmark \\ \hline
\textbf{INO (Ours)} & \cmark & \cmark & \cmark &  \cmark \\
\hline
\end{tabular}}
}
\end{center}
\label{table:method-comparision}
\vspace{-5mm}
\end{table}

%% file: section/2_related_work.tex
\noindent\textbf{Vision Transformers for Dense Prediction.}
Most previous VOS methods~\cite{Dense_araslanov2021dense,VRW_jabri2020space,MAST_lai2020mast,TimeCycle_wang2019learning,Rethinking_xu2021rethinking} adopt convolutional neural networks (CNNs) as the architectures to learn the visual representations.
Recently, the  development~\cite{ViT_dosovitskiy2020image,DeiT_touvron2021training} of Vision Transformer (ViT) architecture shows overwhelming success over CNNs.
Beyond the simple image classification tasks~\cite{SwinTrans_liu2021swin}, researchers have successfully adopted the ViT architecture in several dense prediction tasks, \textit{e.g.},  object detection~\cite{DERT_carion2020end}, semantic segmentation~\cite{VisTR_wang2021end}, \textit{etc.}
However, these works mainly focus on still images and study the fully-supervised settings.
In contrast, our INO is specially designed to solve the video object segmentation via self-supervised training.
We exploit the architecture privilege of ViT by using the naturally designed high-level class tokens and fine-grained patch tokens. 

\noindent\textbf{VOS via Self-supervised Learning.}
Some previous approaches~\cite{VOS2_liang2021rethinking,VOS1_miao2020memory,STM_oh2019video,VOS3_yang2019going,CFBI_yang2021collaborative} learn visual representation fully supervised by annotated datasets with pixel-wise labels.
However, these fully-supervised VOS methods are highly restricted by the diversity of annotated categories and the scale of the annotated dataset.
In this work, we explore a more promising and efficient way which is to leverage self-supervised learning~\cite{DINO_caron2021emerging,MOCO_he2020momentum} to conduct VOS. Previous self-supervised works either only consider the high-level global representations \cite{Rethinking_xu2021rethinking,VRW_jabri2020space}, or directly perform the contrastive learning on the fine-grained pixel-wise features \cite{Dense_araslanov2021dense}. 
Benefiting from the flexibility of vision transformer-based architecture, we optimize both high-level semantics on class tokens and fine-grained semantics on patch tokens, which brings better correspondence learning.

\begin{figure*}[t]
\small
\centering
\includegraphics[width=\textwidth]{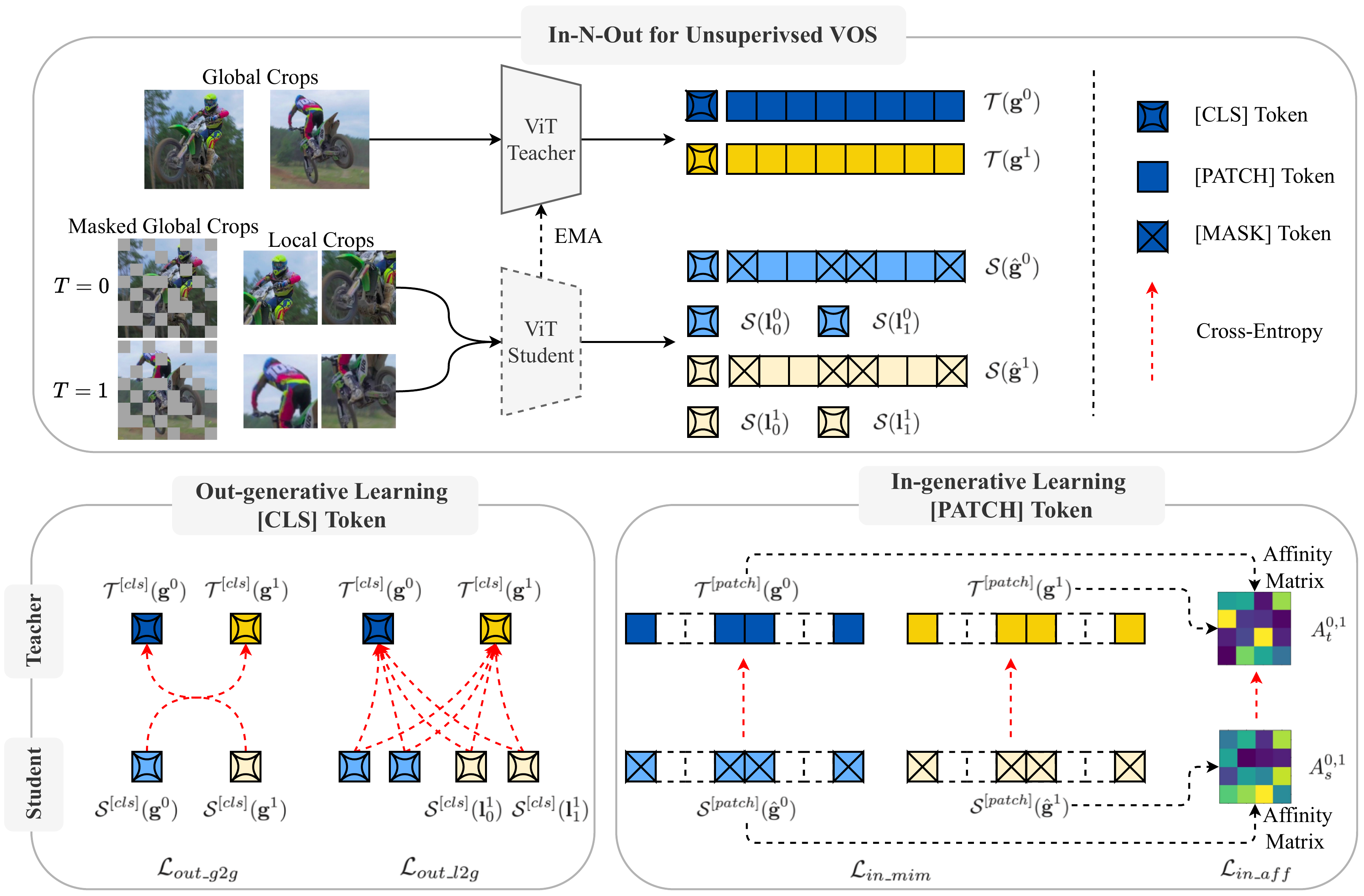}
\caption{\textbf{Overview of INO framework.} For illustration, we assume that the video length $L=2$, and for each frame we crop $1$ global crop and $2$ local crops. 
 (i) \emph{Out-generative learning} is the \emph{out-view} mapping between high-level class tokens, which is composed of $\mathcal{L}_{out \_ g2g}$ and $\mathcal{L}_{out \_ l2g}$. Concisely, $\mathcal{L}_{out \_ g2g}$ is cross-frame calculated between global crop outputs, while $\mathcal{L}_{out \_ l2g}$ is all the possible local-to-global mappings. (ii) \emph{In-generative learning} is the \emph{in-view} recovery on fine-grained patch tokens of global crops, which is composed of $\mathcal{L}_{in \_ mim}$ and  $\mathcal{L}_{in \_ aff}$. In detail, $\mathcal{L}_{in \_ mim}$ is the mapping between mask tokens and the corresponding teacher patch tokens, while $\mathcal{L}_{in \_ aff}$ improves the inter-frame correspondence via bootstrapping.  }
\label{fig_framework}
\vspace{-4mm}
\end{figure*}

\noindent\textbf{From Contrastive Learning to Generative Learning.}
Recently, contrastive learning~\cite{BYOL_grill2020bootstrap,MOCO_he2020momentum} has been popular for self-supervised learning, which models image similarity and dissimilarity between different views.
Although migrating these methods on VOS tasks~\cite{Dense_araslanov2021dense,VRW_jabri2020space,Rethinking_xu2021rethinking, R3_added3_jeon2021mining, R3_added4_zhao2021modelling} has achieved a preliminary success, the contrastive-based methods strongly depend on data augmentation~\cite{VRW_jabri2020space}, and may also suffer the scalability problem~\cite{Dense_araslanov2021dense,VRW_jabri2020space}.
Inspired by masked language modeling~\cite{BERT_devlin2018bert} in NLP, recent attempts~\cite{BEiT_bao2021beit,MAE_he2021masked} learn representations by masked image modeling.
However, these methods either reconstruct the original images in original pixels~\cite{MAE_he2021masked}, or predict the quantized discrete tokens~\cite{BEiT_bao2021beit}.
As illustrated in Fig.~\ref{fig:idea-illustration}, we discover more abundant supervised signals directly in feature embedding space, by recovering masked patch tokens (in-generative) and imagining global information (out-generative).
To the best of our knowledge, we make the first attempt to address self-supervised learning for VOS from a fully generative learning perspective  based on the idea of masked image modeling.

%% file: section/3_method.tex
\noindent\textbf{Overview.} Unsupervised learning for video object segmentation targets on achieving semantically discriminative representations via training on unlabeled videos.
During the inference stage, given the mask annotation of the first frame, the label propagation is performed via the correspondence (similarity) between the extracted feature maps \cite{Dense_araslanov2021dense,VRW_jabri2020space,MAST_lai2020mast,Rethinking_xu2021rethinking}.
As illustrated in Fig.~\ref{fig_framework}, we introduce the proposed In-N-Out generative learning framework for the unsupervised learning for VOS, which is composed of out-generative learning on high-level class tokens and in-generative learning on fine-grained patch tokens.
In \S~\ref{sec:framework} we first introduce the generic teacher-student framework.
\S~\ref{sec:out-generative} demonstrates the proposed out-generative learning targets on mining high-level semantic consistency via the imagination between augmented crops. 
\S~\ref{sec:in-generative} introduces the proposed in-generative learning which focuses on achieving fine-grained semantics via the recovery of the corrupted semantic structure.
We \label{sec:pipeline} present the brief training and inference pipeline of the proposed INO in \S~\ref{sec:pipeline}.



\subsection{Generative Learning Framework}\label{sec:framework}
We use the commonly used teacher-student framework \cite{DINO_caron2021emerging,iBOT_zhou2021ibot} for self-supervised learning.
The teacher $\mathcal{T}$ and student $\mathcal{S}$ share the same architecture which includes a backbone (\textit{e.g.}, ViT \cite{ViT_dosovitskiy2020image,DeiT_touvron2021training}) and a projection head.
Without loss of generality, we directly adopt the original ViT implementation.
The improvement of the backbone model is not the focus of this paper.
The teacher parameters are the Exponentially Moving Average (EMA) of the student parameters. To avoid collapse,  a stop-gradient operator is applied to the teacher, and the teacher output is centered by the computed batch mean \cite{DINO_caron2021emerging}. Then, each network outputs (including both class tokens and patch tokens) are normalized with a temperature softmax to get the final categorical distributions  \cite{iBOT_zhou2021ibot}. The output categorical distributions of the teacher are taken as the generation target of the student output and their similarity is measured with a standard cross-entropy loss. 
Notably, the class tokens mainly capture the high-level semantic information, while the patch tokens focus more on the fine-grained details. Therefore, we choose class tokens for out-generative learning and patch tokens for in-generative learning, respectively.


\subsection{Out-generative Learning}\label{sec:out-generative}

Given the $i$-th frame in a video clip of length $L$,
we first achieve one global crop $\textbf{g}^{i}$ and $M$ local crops $\{ \textbf{l}^{ij} \}_{j=1}^M$ via \textit{random resized cropping} under different scales.
Then, \textit{flipping} and \textit{color jitter} are randomly applied for each crop. 
The augmented crops from frames of the same video clip may be quite different in low-level vision due to the performed augmentation and the motion nature of videos.
However, since they are from the same semantic scene, they share similar high-level semantic meaning. 
We intend to leverage such high-level semantic consistency as the supervision signal for out-generative learning, which can be further divided into global-to-global and local-to-global generation, as illustrated in the lower-left part of Fig. \ref{fig_framework}. 

\noindent\textbf{Temporal Consistency via Global-to-global Generation.}
Considering the consecutive nature of video sequences, the global crops from different frames of a certain sequence may share most of the semantic objects when the scale range for cropping is large. In this case, the most salient difference may come from the motion of objects, which should be recognized for better segmentation label propagation, \textit{e.g.}, the person with different poses in different frames should be recognized as the same person. Therefore, we propose to perform generation between these inter-frame global crops to capture such temporal semantic consistency in the presence of motion. 

Specifically, for a sequence of length $L$, we empirically split the sequence into halves and then zip them as frame pairs $ \mathcal{P} = \{(n, L/2 + n ) \}_{n=1}^{L/2}$. Taking $L=6$ as an example, the formulated pairs are $\{ (1,4), (2,5), (3,6) \}$. 
Given a frame pair $(i_{1},i_{2}) \in \mathcal{P} $, all the global crops $\{ \textbf{g}^{i_1}, \textbf{g}^{i_2} \}$ are sent to the student  together with the teacher, and the global-to-global out-generative learning is
performed as:
\begin{equation}
\mathcal{L}_{out \_ g2g} = \frac{1}{|\mathcal{P} |} \sum_{(i_1,i_2) \in \mathcal{P}} \sum_{\textbf{t} \in \mathbbm{T}  } \sum_{\textbf{s} \in \mathbbm{S} \backslash \{ \textbf{t} \}  } - \mathcal{T}^{[cls]}(\textbf{t})^{T} log \mathcal{S}^{[cls]}(\textbf{s}),
 \label{eq:out_g2g}
\end{equation}
where $ \mathbbm{T} = \{ \textbf{g}^{i_1}, \textbf{g}^{i_2} \} $, $ \mathbbm{S} = \{ \textbf{g}^{i_1}, \textbf{g}^{i_2} \}  $.

\noindent\textbf{Semantic Correlation via Local-to-global Generation.}
Compared with the global-to-global generation, local-to-global generation solves a harder task, which is to imagine the complete scene given limited information from random semantic fragments. In this scenario, not only motion consistency, but also the inference of the high-level semantic correlation is required.
For example, given the fragment of a motorcross and a part of the human leg, the model is required to infer that ``a rider is riding a motorcross'', as illustrated in Fig. \ref{fig:idea-illustration}. Such high-level inference helps the model to capture more robust and complete semantics, which is beneficial to the stable propagation of masks. 

Specifically, for frame pair $(i_{1},i_{2}) \in \mathcal{P} $, all the local crops are sent to the student and calculated with the previous teacher outputs of global crops. Similar to Eq \ref{eq:out_g2g}, the local-to-global generative learning is performed as:
\begin{equation}
\label{eq:out_l2g}
\mathcal{L}_{out \_ l2g} =   \frac{1}{|\mathcal{P} |} \sum_{(i_1,i_2) \in \mathcal{P}} \sum_{\textbf{t} \in \mathbbm{T}  } \sum_{\textbf{s} \in \mathbbm{S}  } - \mathcal{T}^{[cls]}(\textbf{t})^{T} log \mathcal{S}^{[cls]}(\textbf{s}),
\end{equation}
where $ \mathbbm{T} = \{ \textbf{g}^{i_1}, \textbf{g}^{i_2} \} $, $ \mathbbm{S} = \{ \textbf{l}^{{i_1}j}, \textbf{l}^{{i_2}j} \}_{j=1}^{M} $.

\input{table/DAVIS_SOTA}
\input{table/YTVOS_SOTA}

\subsection{In-generative Learning}
The out-generative learning leverages the class tokens to capture the high-level semantic consistency, however, the dense  segmentation task requires more fine-grained semantic information to precisely capture the correspondence.
Therefore, we introduce the in-generative learning on patch tokens of global crops, which is composed of intra-frame masked image modeling and inter-frame affinity consistency, as shown in the lower-right part of Fig. \ref{fig_framework}. 

\noindent\textbf{Intra-frame Masked Image Modeling.}\label{sec:in-generative}
Inspired by \cite{BEiT_bao2021beit,iBOT_zhou2021ibot}, we first perform blockwise masking on the input global crops of the \emph{student} module (as illustrated in Fig. \ref{fig_framework}). Specifically, assuming that a global crop from the $i$-th frame $\textbf{g}^i$ is split into $P$ tokens $\textbf{g}^{i} = \{ g_{j}^i \}_{j=1}^P$, we then randomly choose a ratio $r$ as the proportion of masked tokens and get a random mask $ \textbf{m} \in \{0, 1\}^{P}$.
According to $\textbf{m}$, the picked $K = P \cdot r$ tokens $\{ g_{i} | m_i = 1 \}$ are replaced with a global learnable mask token 
and get a corrupted token sequence $\hat{\textbf{g}}^i$. After that, $\hat{\textbf{g}}^i$ is sent to the following attention module and prediction head for categorical distributions output. For the \emph{teacher} module, the original unmasked token sequence $\textbf{g}^i$ is kept as input. Finally, similar as Eq. \ref{eq:out_l2g}, we take the corresponding teacher outputs of the masked tokens as the generation target and 
calculate the cross-entropy between them for each global crop in a video clip:

\begin{equation}
\label{eq:in_mim}
\mathcal{L}_{in \_ mim} = \frac{1}{L} \sum_{i=1}^{L} \sum_{j=1}^{P} m_{j} \cdot -\mathcal{T}^{[patch]}_{j}(\textbf{g}^i)^{T} log \mathcal{S}^{[patch]}_{j}(\hat{\textbf{g}}^i) .
\end{equation}

Intuitively, with Eq. \ref{eq:in_mim}, we force the model to generate the corrupted patches based on the limited semantic information from the reserved patches. Such supervision forces the model to achieve the complete and fine-grained semantics, which is important for recognizing the gradually appearing parts given incomplete labels in the first frame. For example, the figure of the person may be incomplete (\textit{i.e.}, corrupted) at the first time of appearance due to the occlusion or limited shooting angle. The model should possess the ability to recognize the rest part of a person which may gradually appear in the following frames, and this is in line with the target of $\mathcal{L}_{in \_ mim}$.

\noindent\textbf{Inter-frame Affinity Consistency.}
$\mathcal{L}_{in \_ mim}$ explores the fine-grained semantic information spatially for each global crop, however, the temporal fine-grained semantic consistency between frames should also be considered, therefore, we propose the affinity consistency constrain as follows. 

Given the $i$-th global crop in a video sequence of length $L$, we represent the $l_2$-normalized $d$-dimensional distribution matrix of the masked tokens from the teacher module  as $Q^i_{t} \in \mathbbm{R}^{K \times d}$. Similarly, the corresponding distribution matrix for the student module is represented as $Q^i_{s} \in \mathbbm{R}^{K \times d}$. Then, the affinity matrix from timestep $i$ to $i+1$ for the teacher outputs is calculated as:
\begin{equation}
\label{eq:affinitymatrix}
A_{t}^{i,i+1} = softmax({(Q^i_t} {Q^{i+1}_t}^T)/ \tau_{t}),
\end{equation}
where $A_{t}^{i,i+1} \in \mathbbm{R}^{K \times K}$ and $\tau_{t}$ is the temperature. Similarly, the corresponding affinity matrix and temperature for the student outputs are $A_{s}^{i,i+1}$ and $\tau_{s}$, respectively. 
Then, we calculate the cross-entropy between these two affinity matrices as follows:
\begin{equation}
\label{eq:in_aff}
\mathcal{L}_{in\_aff} = \frac{1}{L-1} \sum_{i=1}^{L-1} \sum_{j=1}^{K} - {A_{t}^{i,i+1} [j,:]}^T log A_{s}^{i,i+1} [j,:] ,
\end{equation}
where $A_{s}^{i,i+1} [j,:]$ represents for the $j$-th row vector of the matrix,
which is the softmax normalized cosine similarity between the $j$-th mask token of frame $i$ and all the $K$ masked tokens of frame $i+1$, \textit{i.e.}, the correspondence. Intuitively, we intend to learn the fine-grained temporal correspondence via bootstrapping, which is beneficial for the ultimate goal of propagating segmentation labels. 





\subsection{Training \& Inference Pipeline}\label{sec:pipeline}
In this subsection, we introduce the whole pipeline for INO to achieve unsupervised learning for VOS.

During the training stage, we start by training the ViT backbone with the raw data from video recognition datasets Kinetics-400 \cite{Kinetics-400_2_kay2017kinetics} and Charades \cite{Charades_sigurdsson2016hollywood} without using any human annotation labels.
Our network is trained in a self-supervised manner with learning objectives:
\begin{equation}
\label{eq:all}
\mathcal{L}_{INO} = \mathcal{L}_{out \_ g2g} + \mathcal{L}_{out \_ l2g} + \mathcal{L}_{in \_ mim} + \mathcal{L}_{in \_ aff} . 
\end{equation}
To maintain simplicity, here we treat all these terms with equal contributions.
Once the backbone model is trained, we can evaluate directly on the DAVIS-2017 \texttt{val}   \cite{DAVIS2017_pont20172017} and YouTube-VOS 2018 \texttt{val} \cite{YouTube-Vos_xu2018youtube} without fine-tuning. 

During inference, the segmentation label of the first frame is provided and then propagated toward the following frames based on the similarity between extracted feature maps. For fair comparison, we use the same label propagation strategy as \cite{Dense_araslanov2021dense,DINO_caron2021emerging,VRW_jabri2020space,Rethinking_xu2021rethinking}, detailed in the supplementary material.


%% file: table/DAVIS_SOTA.tex
\begin{table*}[t]
\small
\caption{\textbf{Comparisons with state-of-the-art methods on DAVIS-2017 \texttt{val}.} RN-18 and ViT-S/8 represent for ResNet-18 and Vit-Small with a patch size of 8, separately. ``\dag'' means using 2 times larger resolution for inference. We also report the number of video sequences (N) and the total video duration (T) for each dataset.  }
\vspace{-2mm}
\begin{center}
 \setlength{\tabcolsep}{5.2mm}{
\begin{tabular}{l|ccc|ccccc}
\rowcolor{LightGray}
\hline
\textbf{Method}       & \textbf{Arch} & \textbf{Dataset} & \textbf{N/T} & \textbf{\JFm}  & \textbf{\Jm}   & \textbf{\Jr}   & \textbf{\Fm}   & \textbf{\Fr}   \\ \hline \hline
TimeCycle \cite{TimeCycle_wang2019learning}    & RN-18                     & VLOG                  & 114K / 344h  & -             & 40.1          & -             & 38.3          & -             \\
TimeCycle \cite{TimeCycle_wang2019learning}    & RN-50                     & VLOG                  & 114K / 344h  & -             & 41.9          & -             & 39.4          & -             \\ \hline
CorrFlow\dag \cite{CorrFlow_lai2019self}     & RN-18                     & Kinetics              & 300K / 833h  & 49.5          & 47.7          & 53.2          & 51.3          & 56.5          \\
CorrFlow\dag \cite{CorrFlow_lai2019self}     & RN-18                     & OxUvA                & 366 / 14h    & 50.3          & 48.4          & 53.2          & 52.2          & 56.0          \\ \hline
ContCorr \cite{ContrastCorr_wang2020contrastive} & RN-18                       & TrackingNet           & 30K / 140h   & 63.0          & 60.5          & -             & 65.5          & -             \\ \hline
MAST\dag \cite{MAST_lai2020mast}         & RN-18                  & OxUvA                 & 366 / 14h    & 63.7          & 61.2          & 73.2          & 66.3          & 78.3          \\
MAST\dag \cite{MAST_lai2020mast}         & RN-18                  & YT-VOS                & 4.5K / 5h    & 65.5          & 63.3          & 73.2          & 67.6          & 77.7          \\ \hline
CRW \cite{VRW_jabri2020space}          & RN-18                   & Kinetics              & 300K / 833h  & 67.6          & 64.8          & 76.1          & 70.2          & 82.1          \\ \hline
JSTG \cite{R3_added4_zhao2021modelling}  & RN-18                   & Kinetics              & 300K / 833h  & 68.7          & 65.8          & 77.7           & 71.6          & 84.3          \\ \hline
DUL \cite{Dense_araslanov2021dense}          & RN-18                    & OxUvA               & 366 / 14h    & 65.3          & 63.4          & 76.1          & 67.2          & 79.7          \\
DUL \cite{Dense_araslanov2021dense}          & RN-18                    & Kinetics              & 300K / 833h  & 68.7          & 66.7          & 81.4          & 70.7          & 84.1          \\
DUL \cite{Dense_araslanov2021dense}          & RN-18                       & YT-VOS                & 4.5K / 5h    & 69.3          & 67.1          & 81.2          & 71.6          & 84.9          \\
DUL \cite{Dense_araslanov2021dense}          & RN-18                    & TrackingNet           & 30K / 140h   & 69.4          & 67.1          & 80.9          & 71.7          & 84.8          \\ \hline
VFS \cite{Rethinking_xu2021rethinking}          & RN-18                  & Kinetics              & 300K / 833h  & 67.6          & 64.8          & -             & 70.2          & -             \\
VFS \cite{Rethinking_xu2021rethinking}          & RN-50                 & Kinetics              & 300K / 833h  & 69.4          & 66.7          & -             & 72.0          & -             \\ \hline
\textbf{INO (Ours)}    & ViT-S/8                 & Charades              & 10K / 82h    & 67.0          & 63.7          & 72.7          & 70.4          & 82.9          \\
\textbf{INO (Ours)}    & ViT-S/8                 & Kinetics              & 300K / 833h  & \textbf{72.5} & \textbf{68.7} & \textbf{82.0} & \textbf{76.3} & \textbf{89.0} \\ \hline
\end{tabular}}
\end{center}
\label{table:DAVIS_with_SOTA}
\vspace{-4mm}
\end{table*}

%% file: table/YTVOS_SOTA.tex
\begin{table}[!t]
\small
\caption{ \textbf{Comparisons with state-of-the-art methods on YouTube-VOS 2018 \texttt{val} benchmark}. The object classes in the \texttt{val} set are partly overlapped with the \emph{training} set, therefore the performance is distinguished as ``seen'' and ``unseen'' categories. 
All results are evaluated and reported through the online testing server~\cite{YouTube-Vos_xu2018youtube}.}
\vspace{-2mm}
\begin{center}

\begin{tabular}{l|cccccc}
\hline
\rowcolor{LightGray}
 &  &  & \multicolumn{2}{c}{\textbf{Seen}} & \multicolumn{2}{c}{\textbf{Unseen}} \\\cline{4-7}
\rowcolor{LightGray}
\multirow{-2}{*}{\textbf{Method}}   & \multirow{-2}{*}{\textbf{Dataset}}                               &   \multirow{-2}{*}{\textbf{Mean}}                            & \textbf{\Jm}     & \textbf{\Fm}     & \textbf{\Jm}      & \textbf{\Fm}      \\ \hline \hline
Colorize \cite{Colorize_vondrick2018tracking}                                         & Kinetics                                           & 38.9                           & 43.1            & 38.6            & 36.6             & 37.4             \\ 
CorrFlow\dag \cite{CorrFlow_lai2019self}                                        & OxUvA                                              & 46.6                           & 50.6            & 46.6            & 43.8             & 45.6             \\ 
MAST\dag \cite{MAST_lai2020mast}                                           & YT-VOS                                             & 64.2                           & 63.2            & 64.9            & 60.3             & 67.7             \\ 
CRW \cite{VRW_jabri2020space}                                         & Kinetics                                           & 69.9                           & 68.7            & 70.2            & 65.4             & 75.2             \\ 
DUL \cite{Dense_araslanov2021dense}                                            & YT-VOS                                           & 69.9                           & 69.6            & 71.3            & 65.0    & 73.5             \\
DUL \cite{Dense_araslanov2021dense}                                         & Kinetics                                           & 70.6                           & 69.9            & 71.3            & \textbf{66.5}    & 74.8             \\
DUL \cite{Dense_araslanov2021dense}                                        & TrackingNet                                        & 70.7                           & 70.2            & 71.9            & 66.3             & 74.5             \\ \hline
\textbf{INO (Ours)}                            & Kinetics                                           & \textbf{71.3}                  & \textbf{70.7}   & \textbf{73.2}   & 65.6             & \textbf{75.6}    \\ \hline
\end{tabular}
\end{center}
\label{table:YTVOS_with_SOTA}
\vspace{-4mm}
\end{table}

%% file: section/4_experiment.tex
\input{figure/DAVIS_VIS}


\input{figure/YTVOS_VIS}

\subsection{Experimental Settings}
\textbf{Datasets.}
In order to validate the scalability of the proposed INO, we conduct experiments on two large-scale datasets, including Charades \cite{Charades_sigurdsson2016hollywood} and Kinetics-400 \cite{Kinetics-400_2_kay2017kinetics}. 
Charades \cite{Charades_sigurdsson2016hollywood} dataset spans 9848 videos with an average length of 30s and records the causal everyday activities at home.
Kinetics \cite{Kinetics-400_2_kay2017kinetics} dataset contains significantly more video sequences (around 230K videos with 10s per video on average). 
Note that we directly use raw data from the video dataset, \textit{no} human annotations are involved during the training process.

\noindent\textbf{Evaluation Metrics.}
To verify the generalization ability of INO, we benchmark on two challenging video object segmentation benchmarks: DAVIS-2017 \texttt{val} \cite{DAVIS2017_pont20172017} and YouTube-VOS 2018 \texttt{val} \cite{YouTube-Vos_xu2018youtube}. DAVIS-2017 \texttt{val} contains $30$ videos in 480p, and  YouTube-VOS 2018 \texttt{val} spans 474 videos, and over $90\%$ of them are in 720p. Following previous works~\cite{Dense_araslanov2021dense,Rethinking_xu2021rethinking,VRW_jabri2020space}, we use region similarity ($\mathcal{J}$) and contour accuracy ($\mathcal{F}$) as the evaluation metric \cite{JF_perazzi2016benchmark}.  

\noindent\textbf{Implementation Details.}
    Our INO framework adopts ViT as the backbone model. For fair comparison with the competing methods \cite{Dense_araslanov2021dense,VRW_jabri2020space,Rethinking_xu2021rethinking} that downsample the input resolution for 8 times during inference, we use the ViT-S/8 configuration by default, which possesses comparable parameters as ResNet-50 \cite{ResNet_he2016deep}.  
    For each iteration, we do in-generative learning with a random probability of $0.5$, and $r$ is randomly sampled from a uniform distribution from $0.1$ to $0.5$.
    The temperature is set as $\tau_{s} = 0.1$ and $\tau_{t} = 0.04$, separately.
    The scale range for global and local crops are $(0.05,0.8)$ and $(0.8, 0.95)$, respectively.
    The global crops are resized to $224 \times 224$ while the local crops are resized to $64 \times 64$. The number of local crops $M$ is set as $8$, and the video length $L$ is set as $4$. 
    We train INO for 25 epochs on both Charades and Kinetics-400 with 4 and 8 V100 (16GB) GPUs, respectively.
    Previous works \cite{Dense_araslanov2021dense,VRW_jabri2020space,Rethinking_xu2021rethinking} use the last-block output of ResNet for training, while the middle-block output (\textit{e.g.}, \emph{res 3} block in \cite{VRW_jabri2020space}) for inference. 
    Similarly, we use the output of the last layer (\textit{i.e.}, the $12$-th layer) of ViT for training and empirically use the output of the $7$-th layer for inference.
    More implementation details can be found in the supplementary material. 

\vspace{-3mm}
\subsection{Quantitative Comparisons with State-of-the-art}

\textbf{Results on DAVIS-2017 benchmark.} 
    In Table \ref{table:DAVIS_with_SOTA}, we report the performance comparisons between our INO and other competing state-of-the-art methods on DAVIS-2017 \texttt{val} benchmark. 
    In a fair comparison where all the methods use the large-scale Kinetics dataset for training, our INO significantly outperforms other state-of-the-art methods by a large margin. 
    For instance, we outperform the second best VFS~\cite{Rethinking_xu2021rethinking} by $3.1\%$ in terms of \JFm ( $72.5\%$ \textit{vs.} $69.4\%$).
    Notably, MAST \cite{MAST_lai2020mast} and CorrFlow \cite{CorrFlow_lai2019self} adopt $2$ times larger image resolution than ours during inference, which leads to larger memory footprint. 
    However, our INO still outperforms MAST~\cite{MAST_lai2020mast} by $7 \%$ on \JFm (72.5\% \textit{vs.} 65.5\%).

\noindent\textbf{Better Scalability.}
Scalability is an important criterion for self-supervised learning methods.
However, as shown in Table~\ref{table:DAVIS_with_SOTA}, the performance of CorrFlow, MAST, and DUL is not positively correlated with the scale of the training dataset. 
And the best performance is not achieved on the largest dataset.
For instance, even though the dataset size is two orders of magnitude smaller ($300k$ \textit{vs.} $4.5k$), the performance of DUL trained with YouTube-VOS is still higher than Kinetics by $0.6 \%$ in \JFm score (68.7 \% \textit{vs.} 69.3 \%). 
Actually, for such methods, the best performance are achieved on the datasets with higher visual quality and cleaner background, \textit{e.g.}, YouTube-VOS \cite{YouTube-Vos_xu2018youtube}, TrackingNet \cite{TrackingNet_Muller_2018_ECCV}, and OxUvA \cite{OxUvA_valmadre2018long}.
This demonstrates that the previous methods are sensitive to the dataset quality and lack of scalability. 
In contrast, our INO exhibits strong scalability and robustness owing to the generative learning paradigm, \textit{e.g.}, the performance boosts by $5.5$\% from Charades to Kinetics.

\noindent\textbf{Results on YouTube-VOS 2018 benchmark.} 
Following~\cite{Dense_araslanov2021dense,MAST_lai2020mast}, we additionally evaluate INO on YouTube-VOS 2018 \texttt{val} dataset in Table~\ref{table:YTVOS_with_SOTA}, which is a more challenging benchmark.
Our INO reaches a new state-of-the-art which improves over DUL \cite{Dense_araslanov2021dense} by  1.3\% in \Fm  of the ``seen'' category and 0.6\% in mean score.
Compared with CRW \cite{VRW_jabri2020space} which is also trained on Kinetics, our INO outperforms by 1.4\% in the mean score. As to MAST \cite{MAST_lai2020mast} which adopts 2 times larger resolution and a more advanced two-stage inference pipeline, \textit{i.e.}, detecting a ROI first and then considering the correspondence bounded by the ROI, our INO still outperforms by a significant margin ($71.3$\% \textit{vs.} $64.2$\%). 
\vspace{-3mm}

\subsection{Qualitative Analysis}
We visualize the mask propagation results in Fig. \ref{fig_DAVIS_vis} and Fig. \ref{fig_YTVOS_vis} for DAVIS-2017 \texttt{val} and YouTube-VOS 2018 \texttt{val}, respectively. 
For better comparison, we also illustrate the results for DUL \cite{Dense_araslanov2021dense} and CRW \cite{VRW_jabri2020space}.
We observe that our INO achieves better qualitative results and is superior in the following aspects:

\noindent\textbf{Robust to Unseen Parts.}
As illustrated in the first ``motorcross-jump'' scenario of Fig. \ref{fig_DAVIS_vis}, the right leg and back view of the rider is unseen in the given label of the first frame.
The contrastive learning based methods CRW \cite{VRW_jabri2020space} and DUL \cite{Dense_araslanov2021dense} fail to track such unseen parts in the following sequences, while our INO tracks them successfully (see frame $15$ and $35$). We attribute this superiority mainly to the in-generative learning objective, which helps capture the complete semantic structure of the corrupted parts. 

\noindent\textbf{Robust to Deformation.} 
Our INO shows better robustness to deformation compared with previous methods.
For instance, in the ``motorcross-jump'' scenario of Fig. \ref{fig_DAVIS_vis}, the back wheel deforms significantly during the motion process, while only our INO can consistently identify the entire back wheel. 
The out-generative learning process fully exploits the semantic consistency between deformed objects from different frames during training. 
Intuitively, this supervised signal improves the robustness of features toward the deformation. 

\noindent\textbf{Less Artifacts.}
During the mask propagation process, the label may shift toward the background which shares a similar pattern with the target object, and this is termed as the ``bleeding'' artifacts in \cite{Dense_araslanov2021dense}. 
As illustrated in the first ``motorcycle'' case in Fig. \ref{fig_YTVOS_vis}, both CRW \cite{VRW_jabri2020space}
and DUL \cite{Dense_araslanov2021dense} suffer this issue in frame 65 and 85, while our INO gives a more complete and stable mask even under a noisy background. 

\noindent\textbf{Fine-grained Correspondence.}
In the second ``hat-trick'' case of Fig. \ref{fig_YTVOS_vis}, we illustrate an extremely hard scenario that requires the ability to precisely capture the fine-grained details. Specifically, a small hat in high motion is thrown by the actor. As the label propagates, CRW \cite{VRW_jabri2020space} lost the hat gradually, and DUL \cite{Dense_araslanov2021dense} shifts to the actor's face immediately, while only our INO tracks the hat consistently and precisely. 


\input{table/Ablation}

\subsection{Ablation Studies}
\label{sec_ablation_study}
We investigate the effectiveness of learning objectives in Table \ref{table:Ablation}, and the influence of training parameters in Fig. \ref{fig_ablation}. The performance trained on Charades and evaluated on DAVIS-2017 \texttt{val} is reported.

\noindent\textbf{Effectiveness of In-N-Out Generative Learning.}
In Table~\ref{table:Ablation}, we investigate the effectiveness of In-N-Out generative learning objectives. We set the out-generative learning between global crops as the baseline configuration and then add each term gradually.
With only out-generative learning between global crops
($\mathcal{L}_{out \_ g2g}$), the model can achieve 45.6\% in \JFm. 
After adding the out-generative learning between local and global crops ($\mathcal{L}_{out \_ l2g}$), the \JFm score boosts for
8.6\%, which shows that the high-level semantic inference from local to global crops can significantly help to improve the performance.
With out-generative learning only, the model achieves a performance of 54.2\%. 
Notably, the performance is further improved significantly by 11.2\% in \JFm score after including the in-generative learning via masked image modeling ($\mathcal{L}_{in \_ mim}$), which shows the importance of fine-grained structural semantic. Finally, the best performance is achieved after $\mathcal{L}_{in \_ aff}$ is introduced, which improves the temporal correspondence via bootstrapping.


\input{figure/Ablation}

\noindent\textbf{Temperature of Affinity Constrain.}
In $\mathcal{L}_{in \_ aff}$, we use $\tau_{s}$ and $\tau_{t}$ as the temperature for student and teacher affinity matrix, separately.
We fix the student temperature $\tau_{s}$ as $0.1$ and change $\tau_{t}$ as $\{0.02, 0.04, 0.06\}$, where smaller $\tau_{t}$ gives sharper target distributions. As illustrated by the yellow one in Fig.~\ref{fig_ablation}, $\tau_{t}=0.04$ gives a moderate sharpness and performs best. 

\noindent\textbf{Sequence Length.} 
We increase the sequence length $L$ as $\{2,4,6\}$, and find that $L=4$ performs best, as illustrated by the orange one in Fig. \ref{fig_ablation}. Longer sequence length brings larger difference between the paired frames, however, too much difference (\textit{e.g.}, the object may disappear in the later frames) incurs the mismatch between the source and target views, which increases the training noise.
Therefore, a moderate sequence length ($L=4$) gives the best performance.

\noindent\textbf{Number of Local Crops.} %
We vary the local crops number $M$ as $\{6, 8, 10 \}$. More local crops bring more radical high-level semantic inference. As illustrated by the green one in Fig. \ref{fig_ablation}, $M=8$ gives the best performance, which is tangibly better than $M=6$ and $M=10$, \textit{e.g.}, $M=10$ leads to a drop in \JFm score by 2.1\%. We infer that too many local crops may lead to overfitting toward the noise, and a moderate $M=8$ is sufficient. 

\noindent\textbf{Scale Threshold for Random Resized Cropping.}
We use the random resized cropping which first samples views with scale range $(0.05, s)$ for local crops and $(s, 0.95)$ for global crops, and then resized them to $64 \times 64$ and $224 \times 224$, respectively. We vary the threshold $s$ as $\{0.4, 0.6, 0.8\}$. Larger $s$ leads to more diverse local crops and more stable global crops containing richer information about the scene. By observing the gray one in Fig. \ref{fig_ablation}, we find that the performance becomes better as $s$ increases from $0.4$ to $0.8$. We infer  that the global crops with rich global information are more helpful to the semantic inference from local to global.

\noindent\textbf{Flipping and Color Jitter.}
\input{table/FC_Ablation}
After obtaining global and local crops via \emph{random resized cropping},  we empirically perform  \emph{flipping} and \emph{color jitter} (F\&C) randomly only for local crops. We investigate  $4$ configurations in Table \ref{table:FC_Ablation}. We observe that operating F\&C on global crops will reduce the performance. For instance, compared with not using F\&C, the \JFm score drops by 3.0\% from 66.6\% to 63.6\% after performing F\&C on global crops. We infer that more stable global crops are preferred for the generation process. Performing F\&C only on local crops provides the best performance on average (67.0\% in \JFm score), therefore we take it as the default configuration.

\input{figure/ALL_badcase}
\subsection{Limitation}
\label{limitation}
We provide failure cases on DAVIS-2017 \texttt{val} and YouTube-VOS 2018 \texttt{val} in Fig. \ref{fig_all_badcase}. We observe that the performance tends to be unsatisfactory under the following challenging scenarios: (i) \emph{Multiple extremely similar instances.} For instance, the two dogs in case 1 of Fig. \ref{fig_all_badcase} are very similar in low-level vision, therefore the labels are confused during the later time steps. (ii) \emph{Long-term and large-area occlusion.} As illustrated by case 2 of Fig. \ref{fig_all_badcase}, the labels tend to lose the object after the long-term and large-area occlusion.
We can observe that there is still room for improvement in the aforementioned challenging cases.






%% file: figure/DAVIS_VIS.tex
\begin{figure*}[t]
\centering
\includegraphics[width=\linewidth]{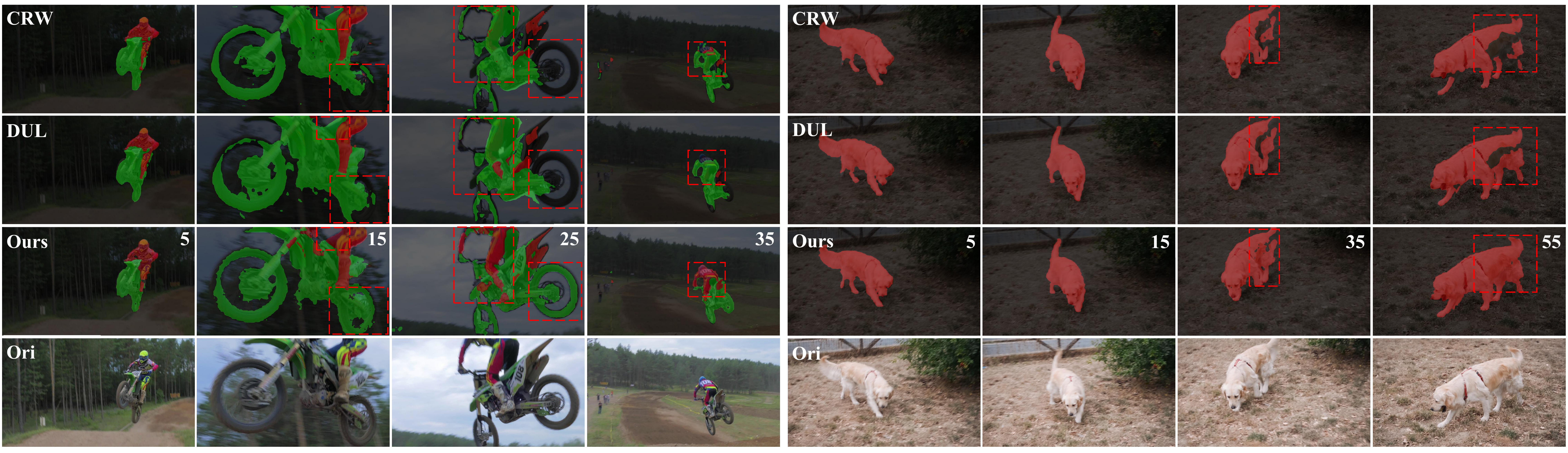}
\vspace{-2mm}
\caption{\textbf{Qualitative comparisons on DAVIS-2017 \texttt{val}.} 
We also display the examples of DUL \cite{Dense_araslanov2021dense} and CRW \cite{VRW_jabri2020space}. 
The frame number is illustrated in the upper-right corner.
We mark the salient parts where our method performs better with red dotted boxes. 
}
\vspace{-3mm}
\label{fig_DAVIS_vis}
\end{figure*}

%% file: figure/YTVOS_VIS.tex
\begin{figure*}[t]
\centering
\includegraphics[width=\linewidth]{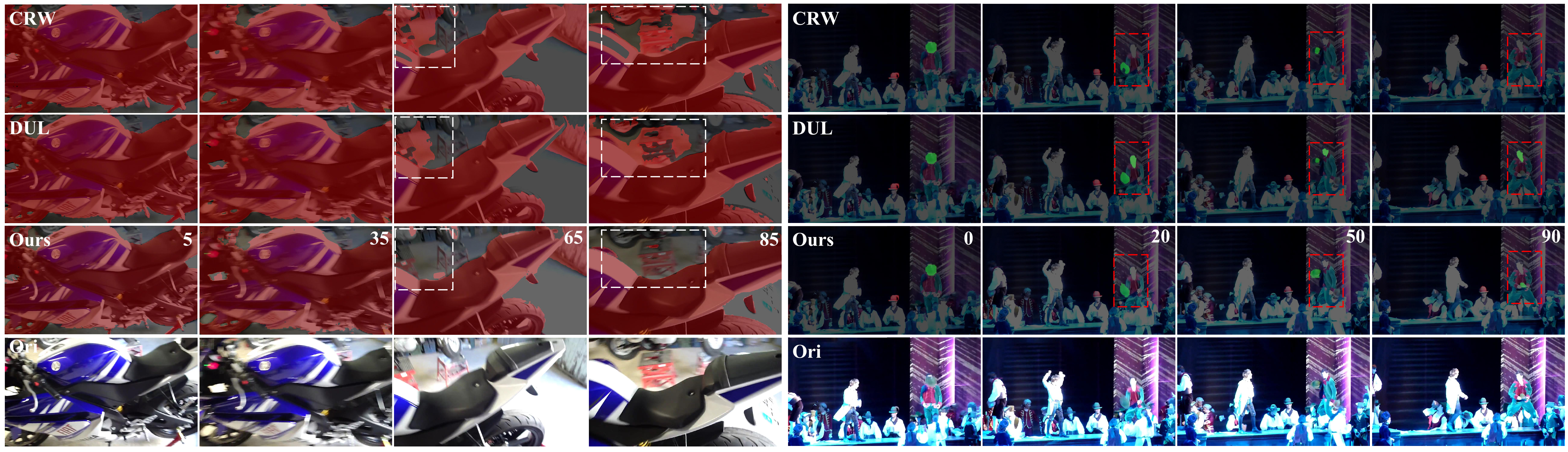}
\vspace{-2mm}
\caption{\textbf{Qualitative comparisons on YouTube-VOS 2018 \texttt{val}}.
Notably, in the second ``hat-trick'' case, we illustrate a hard scenario where the hat is rapidly moving with highly blurry. CRW \cite{VRW_jabri2020space} lost the hat gradually, while DUL \cite{Dense_araslanov2021dense} mistracks toward the human face. Only our INO succeeds to track the hat in motion, which shows that our method can precisely capture the fine-grained correspondence.}
\label{fig_YTVOS_vis}
\vspace{-1mm}
\end{figure*}

%% file: table/Ablation.tex
\begin{table}[t]
\small
\caption{\textbf{Effectiveness of In-N-Out generative learning objectives.} All results are evaluated on DAVIS-2017 \texttt{val} benchmark. }
\vspace{-3mm}
\begin{center}
 \setlength{\tabcolsep}{3mm}{
\begin{tabular}{l|l|ccccc}
\hline
\rowcolor{LightGray}

                  & \textbf{Objective} & \textbf{\JFm}  & \textbf{\Jm}   & \textbf{\Jr}   & \textbf{\Fm}   & \textbf{\Fr}   \\ \hline
\multirow{2}{*}{\rotatebox{90}{Out}} & $\mathcal{L}_{out \_ g2g}$          & 45.6          & 43.4          & 46.3          & 47.7          & 54.8          \\
                  & $+\mathcal{L}_{out \_ l2g}$        & 54.2          & 51.0          & 56.1          & 57.4          & 64.0          \\ \hline
\multirow{2}{*}{\rotatebox{90}{In}} & $+\mathcal{L}_{in \_ mim}$          & 65.4          & 62.4          & 71.6          & 68.3          & 80.1          \\
                  & $+\mathcal{L}_{in \_ aff}$           & \textbf{67.0} & \textbf{63.7} & \textbf{72.7} & \textbf{70.4} & \textbf{82.9} \\ \hline
\end{tabular}}
\end{center}
\label{table:Ablation}
\vspace{-4mm}
\end{table}

%% file: figure/Ablation.tex
\begin{figure}[t]
\small
\centering
\includegraphics[width=0.9\linewidth]{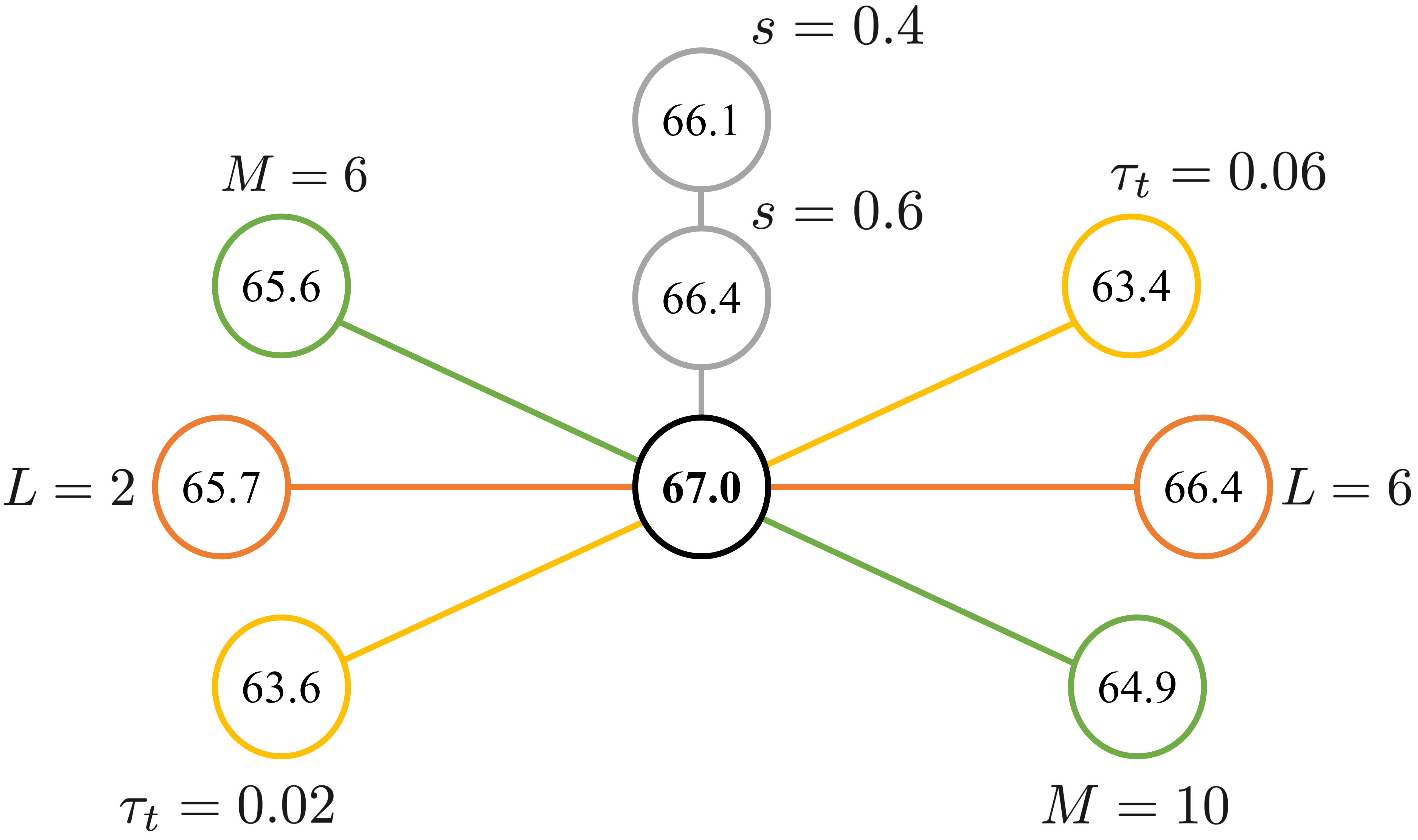}
\caption{\textbf{Ablation study of training parameters.} We report the \JFm  score evaluated on DAVIS-2017 \texttt{val}. Our baseline configuration (the centered one) is: $M=8,L=4,\tau_{t}=0.04,s=0.8$. }
\label{fig_ablation}
\vspace{-2mm}
\end{figure}

%% file: table/FC_Ablation.tex
\begin{table}[t]
\small
\caption{Influence of different configurations of operating \emph{flipping} and \emph{color jitter} (F\&C). 
}
\vspace{-1mm}
\begin{center}
 \setlength{\tabcolsep}{2mm}{
\begin{tabular}{c|c|c|ccccc}
\hline
\rowcolor{LightGray}
\multicolumn{1}{l|}{\textbf{}} & \multicolumn{1}{l|}{\textbf{Global?}} & \multicolumn{1}{l|}{\textbf{Local?}} & \multicolumn{1}{c}{\JFm}            & \multicolumn{1}{c}{\Jm}             & \multicolumn{1}{c}{\Jr}             & \multicolumn{1}{c}{\Fm}             & \multicolumn{1}{c}{\Fr}             \\ \hline
\multirow{4}{*}{F\&C} & -                            & -                       & \multicolumn{1}{c}{66.6} & \multicolumn{1}{c}{63.5} & \multicolumn{1}{c}{\textbf{74.7}} & \multicolumn{1}{c}{69.7} & \multicolumn{1}{c}{82.4} \\ 
                               & \cmark                                 & -                                    & 63.6                              & 60.8                              & 69.5                              & 66.5                              & 77.9                              \\ 
                               & -                                     & \cmark                                & \textbf{67.0}                     & \textbf{63.7}                     & 72.7                              & \textbf{70.4}                     & \textbf{82.9}                     \\ 
                               & \cmark                                 & \cmark                                & 66.2                              & 63.3                              & 73.3                              & 69.1                              & 82.1                              \\ \hline
\end{tabular}}
\end{center}
\label{table:FC_Ablation}
\vspace{-4mm}
\end{table}

%% file: figure/ALL_badcase.tex
\begin{figure}[t]
\centering
\includegraphics[width=\linewidth]{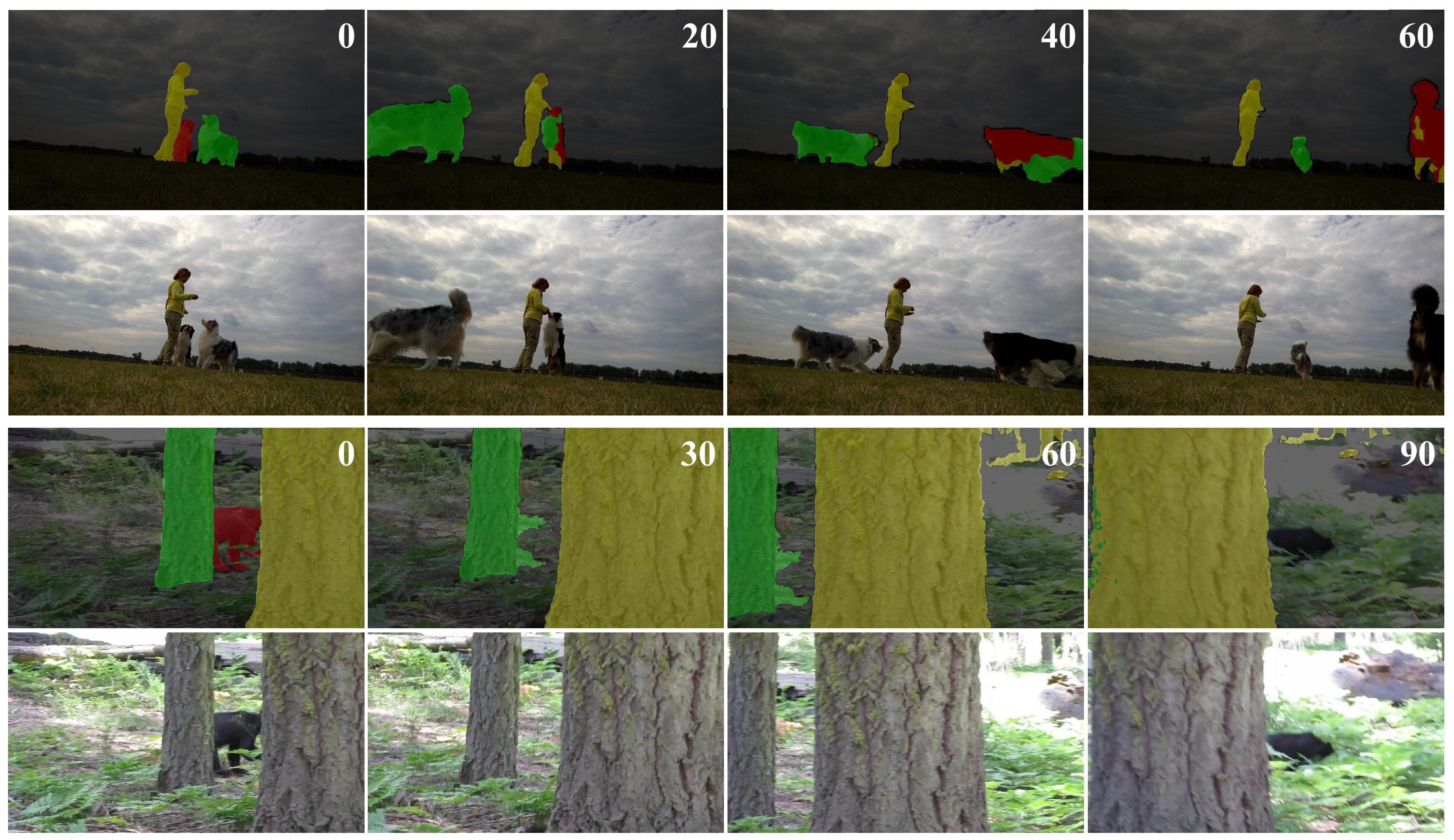}
\vspace{-4mm}
\caption{\textbf{Failure cases on DAVIS-2017 \texttt{val} (upper) and  YouTube-VOS 2018 \texttt{val} (bottom).} 
}
\vspace{-4mm}
\label{fig_all_badcase}
\end{figure}

%% file: section/5_conclusion.tex
In this paper, to tackle the challenging unsupervised learning for VOS, we proposed a simple yet effective framework called In-N-Out (INO) generative learning. 
The proposed INO is a novel fully-generative learning framework via in-view image recovery and out-view image imagination, which integrates both pixel-level and image-level optimization in a unified framework.
Without any annotation data, we effectively learn robust visual representations in a self-supervised manner and achieve the new state-of-the-art performance among unsupervised learning methods for VOS. However, there is still a long way to go considering the analyzed limitations in \S~\ref{limitation}  and the performance gap between the unsupervised and the supervised methods.
Thus, more effective algorithms are still required to alleviate this gap. 
We hope that our efforts will motivate more researchers and ease future research.

%% file: section/appendix.tex
\section{Additional Implementation Details}
We provide more training and inference details in this section. The code will be released upon acceptance.

\subsection{Training Details}
After achieving global and local views via random resized cropping, we empirically perform random flip and random color jitter only for local crops, which leads to more stable global crops.
Learnable position embedding is initialized with the resolution of $64 \times 64$ and resized to the needed resolution (determined by the input resolution) with bicubic interpolation. The architecture of the projection head  is the same as \cite{DINO_caron2021emerging} and its layer number and output dimension are set as $3$ and $4096$, respectively.
We train our network with adamw optimizer \cite{AdamW_loshchilov2018fixing} and the weight decay follows the cosine schedule \cite{CosineScheduler_loshchilov2016sgdr} from 0.04 to 0.4. The momentum of EMA is set as $0.996$. 
We use the same video data loader as the released code of \cite{VRW_jabri2020space} and set the frameskip interval for each clip as $8$. Considering the significant scale difference between these two datasets, we employ different learning rate schedules. Specifically, for Charades, we linearly warmup the learning rate for the first 5 epochs till the base value before decaying with a cosine schedule. The base value is determined by the linear scaling rule \cite{LinearWarmUp_2017accurate}: $lr=0.003 * batchsize * L / 1024$. While for Kinetics-400, we reduce the linear warmup epochs to $2$ and the learning rate is reduced to $lr=0.0003 * batchsize * L / 1024$ since it has significantly more iterations. By default, we train on 4 V100 (16GB) GPUs with a batchsize of $16$ ($4$ clips per GPU) for Charades, and a batchsize of $32$ on 8 V100 (16GB) GPUs for Kinetics-400. 

\subsection{Inference Details}
During inference, the given segmentation labels of the first frame are propagated toward the following frames via the label propagation algorithm. For fair comparison, we employ the same algorithm as previous methods~\cite{DINO_caron2021emerging,Dense_araslanov2021dense,VRW_jabri2020space,Rethinking_xu2021rethinking}. Specifically, given a  $d$-dimensional feature embedding of the target pixel in the current frame, its cosine similarity \textit{w.r.t} the first frame and the $N_c$ previous context frames is calculated. For each frame, restrict attention \cite{MAST_lai2020mast}, which means to only consider the pixels around the target pixel location with a radius of $N_r$, is applied. Then, for all the included reference pixels, the most similar $N_k$ pixels are normalized with a temperature softmax. The final propagated label for the target pixel is the weighted combination of the reference pixel labels.

Same as previous works \cite{Dense_araslanov2021dense,Rethinking_xu2021rethinking,MAST_lai2020mast}, the original image size is kept for inference. The label propagation is calculated via the patch tokens. 
The parameters of the label propagation algorithm on DAVIS-2017 \texttt{val} are $N_k=5,N_c=10,N_r=40$. 
As to YouTube-VOS 2018 \texttt{val}, we adopt longer context ($N_c=20$) and larger radius ($N_r=50$) considering its longer video length in average and larger resolution (720p vs. 480p). 

\section{Additional Ablation Studies}

We supplement more detailed ablation studies in this section. The performance trained on Charades and benchmarked on DAVIS-2017 \texttt{val} is reported. 

\subsection{Ablation of Training Parameters}
\input{figure/sup_ablation}
\subsubsection{Mask Ratio}
We perform masked image modeling at ratio $r$. Similar as \cite{iBOT_zhou2021ibot}, we randomly sample $r$ from a uniform distribution from $r_l$ to $r_h$ at each iteration. We empirically fix $r_l$ as 0.1 and vary $r_h$ as $\{ 0.25, 0.5, 0.75 \}$. As illustrated by the  yellow one in Fig. \ref{fig_sup_ablation}, $r_h=0.5$ performs better than the rest ones. Larger $r_h$ makes the recovery task harder, but may also include more noise. Therefore a moderate $r_h=0.5$ is sufficient. 

\subsubsection{Position Embedding Resolution}
As in \cite{DINO_caron2021emerging}, the learnable position embedding (PE) is initialized with a fixed resolution, and interpolated to the target input resolutions. We tried two initial resolutions based on the used global and local crop size, \textit{i.e.}, $224 \times 224$ and $64 \times 64$. As shown by the blue one in Fig. \ref{fig_sup_ablation}, we find that $64 \times 64$ gives better performance. This is reasonable considering that there are much more local crops (8 per frame) than global clops (1 per frame) during training. Initializing with $64\times64$ resolution ensures that all the parameters in position embedding are fully optimized, while initializing with $224 \times 224$ leads to the downsampling of position embedding in most cases (\textit{i.e.}, when forwarding local crops), therefore the initialized parameters may be optimized in bias. 

\subsubsection{EMA Momentum}
In the generic student-teacher framework, the teacher parameter is the Exponentially Moving Average (EMA) of the student parameters. We investigate the influence of the momentum $m$ by varying it as $\{0.996, 0.9995\}$. As illustrated by the purple one in Fig. \ref{fig_sup_ablation}, we find that $m=0.996$ performs significantly better. 

\subsubsection{Projection Head Layer Number}
We employ the same projection head architecture as \cite{DINO_caron2021emerging}, which is composed of a $N_l$-layer multi-layer perceptron (MLP) and a final fully connected layer with $k$ dimensions output. We test the influence of the layer number $N_l$ via changing it as $\{1,2,3,4\}$. We empirically find that $N_l=1$ gives significantly lower performance, while $N_l=2,3,4$ performs similarly, and $N_l=3$ provides the best performance, as shown by the gray one in Fig. \ref{fig_sup_ablation}. 

\subsubsection{Projection Head Output Dimension}
We investigate the influence of projection head output dimension $k$ via varying it as $\{2048, 4096, 8192\}$ (the orange one in Fig. \ref{fig_sup_ablation}). We observe that $k=4096$ performs best.


\subsection{Ablation of Backbone Architecture}
\input{table/Backbone_Ablation}
We investigate the influence of backbones in Table \ref{table:Backbone_Ablation}. We additionally test ViT-T/8, which shares comparable parameters as ResNet-18 \cite{ResNet_he2016deep}. We observe that the performance can benefit from the increase of backbone parameters, \emph{e.g.}, (67.0\% \emph{vs.} 65.7\% in \JFm  score).

\section{Analysis of Efficiency}
\input{table/efficiency}
 In Table \ref{table:efficiecy}, we compare the efficiency of our INO with the second best method VFS \cite{Rethinking_xu2021rethinking}, which belongs to the image-level optimization using contrastive learning. As illustrated, with the similar training time, our INO employs a more light-weight backbone (21M \emph{vs.} 24M) and requires less GPU memories ($8 \times 16GB$ \emph{vs.} $8 \times 24GB$), while achieves significantly better performance, \emph{e.g.}, 72.5\% \emph{vs.} 69.4\% in \JFm score  (see Table \ref{table:DAVIS_with_SOTA}).

\section{Additional Visualization Examples}
We provide more visualization examples of our INO on DAVIS-2017 \texttt{val} and YouTube-VOS 2018 \texttt{val} in Fig. \ref{fig_DAVIS_vis_sup} and Fig. \ref{fig_YTVOS_vis_sup}, respectively. 
\input{figure/DAVIS_VIS_sup}
\input{figure/YTVOS_VIS_sup}



%% file: figure/sup_ablation.tex
\begin{figure}[h]
\small
\centering
\includegraphics[width=0.8\linewidth]{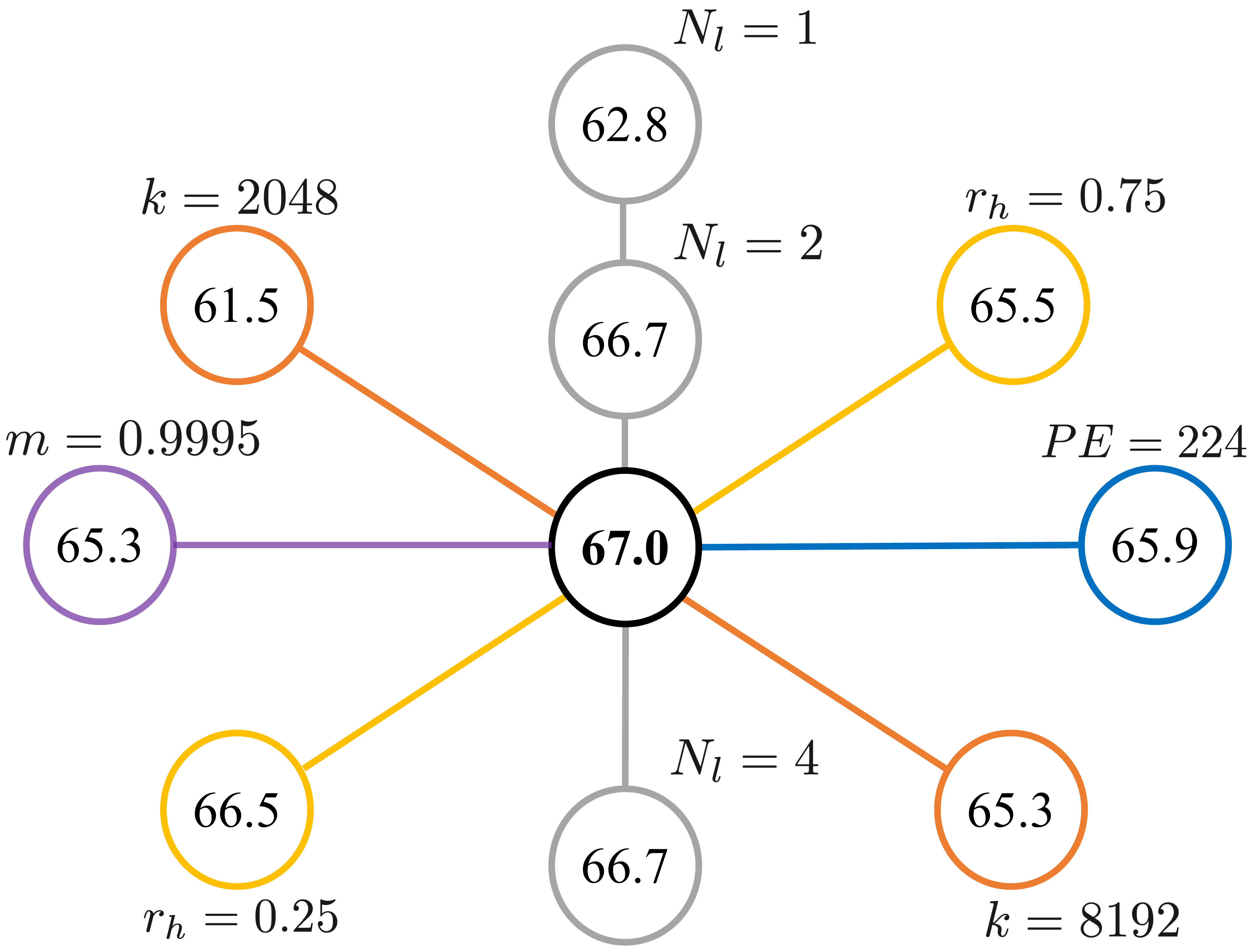}
\caption{\textbf{Additional ablation studies of training parameters.}  We report the \JFm  score evaluated on DAVIS-2017 \texttt{val}. Our baseline configuration (the centered one) is: $m=0.996, k=4096, N_l=3, r_h=0.5, PE=64$. }
\label{fig_sup_ablation}
\vspace{-2mm}
\end{figure}

%% file: table/Backbone_Ablation.tex
\begin{table}[h]
\small
\caption{Influence of backbone architectures. All results are evaluated on DAVIS-2017 \texttt{val} benchmark. }
\vspace{-1mm}
\begin{center}
 \setlength{\tabcolsep}{3.9mm}{
\begin{tabular}{l|ccccc}
\hline
\rowcolor{LightGray}
\textbf{Arch} & \JFm            & \Jm             & \Jr             & \Fm             & \Fr             \\ \hline
ViT-T/8       & 65.7          & 62.8          & \textbf{73.0} & 68.7          & 81.6          \\ 
ViT-S/8       & \textbf{67.0} & \textbf{63.7} & 72.7          & \textbf{70.4} & \textbf{82.9} \\ \hline
\end{tabular}}
\end{center}
\label{table:Backbone_Ablation}
\vspace{-1mm}
\end{table}

%% file: table/efficiency.tex
\begin{table}[h]
\small
\caption{Comparison of efficiency when training on Kinetics-400.}
\vspace{-1mm}
\begin{center}
 \setlength{\tabcolsep}{3.9mm}{
\begin{tabular}{l|lccc}
\hline
\rowcolor{LightGray}
\textbf{Method} & \textbf{Arch} & \textbf{Param} & \textbf{GPUs}    & \textbf{Time} \\ \hline
VFS~\cite{Rethinking_xu2021rethinking}    & RN-50         & 23M              & $8 \times 24GB$ & 1week         \\ 
INO (ours)    & ViT-S/8       & 21M              & $8 \times 16GB$   & 1week         \\ \hline
\end{tabular}}
\end{center}
\label{table:efficiecy}
\vspace{-1mm}
\end{table}

%% file: figure/DAVIS_VIS_sup.tex
\begin{figure*}[t]
\centering
\includegraphics[width=0.9\linewidth]{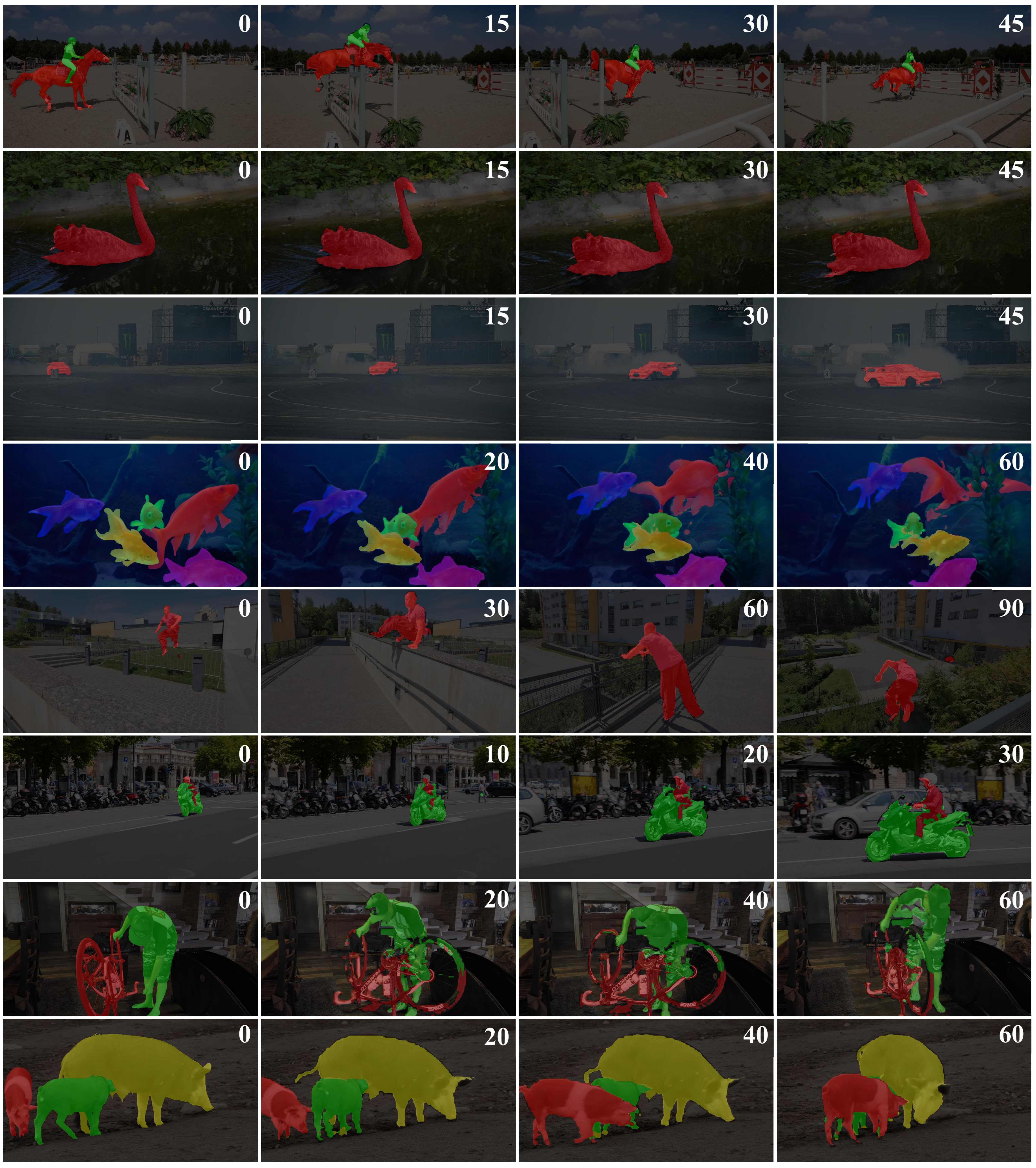}
\vspace{-4mm}
\caption{\textbf{Supplemented visualization examples of our INO on DAVIS-2017 \texttt{val}.} 
}
\vspace{-3mm}
\label{fig_DAVIS_vis_sup}
\end{figure*}

%% file: figure/YTVOS_VIS_sup.tex
\begin{figure*}[t]
\centering
\includegraphics[width=0.9\linewidth]{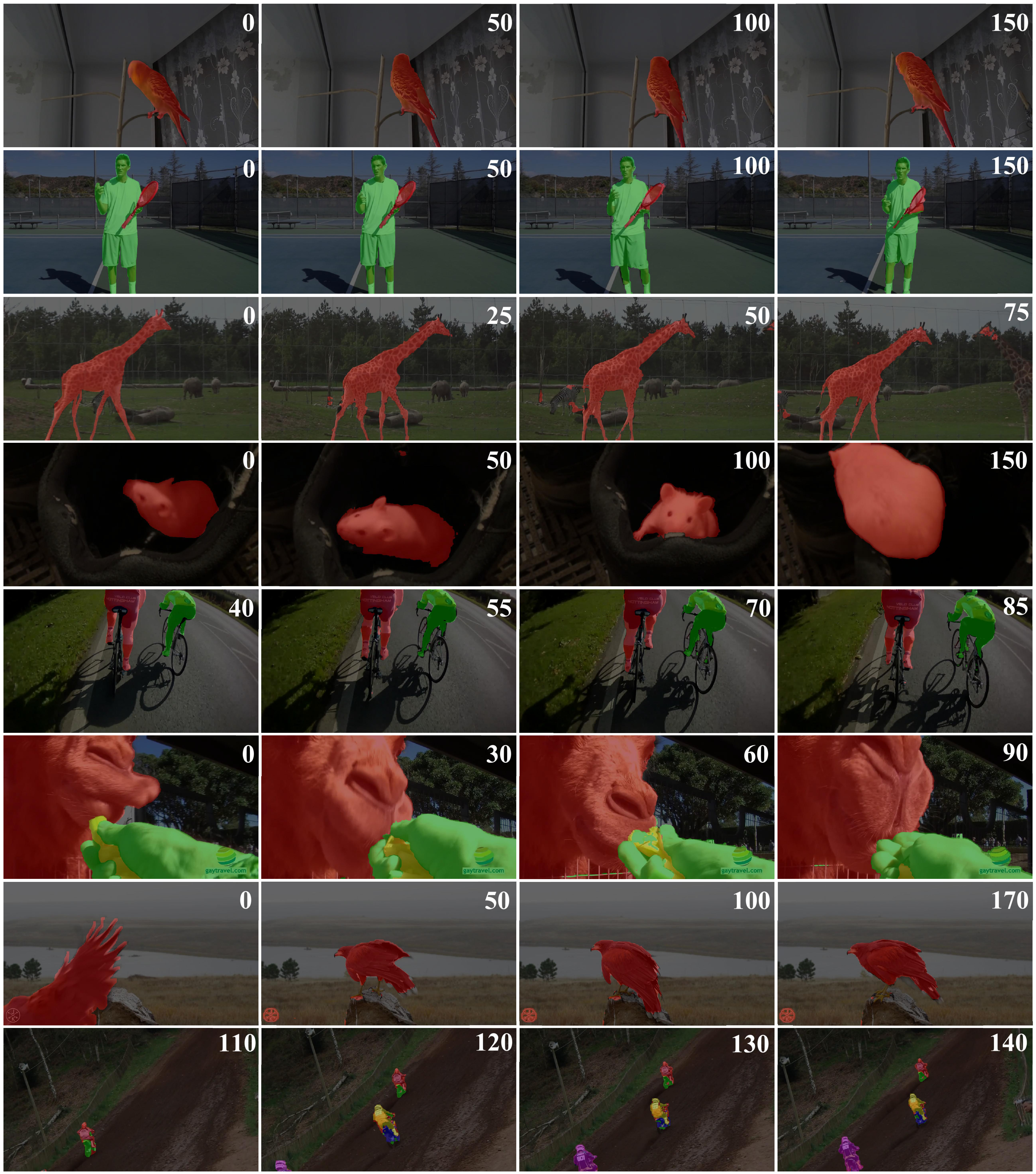}
\vspace{-4mm}
\caption{\textbf{Supplemented visualization examples of our INO on YouTube-VOS 2018 \texttt{val}.} 
}
\label{fig_YTVOS_vis_sup}
\end{figure*}

%% file: sample-sigconf.bbl

\begin{thebibliography}{44}


\ifx \showCODEN    \undefined \def \showCODEN     #1{\unskip}     \fi
\ifx \showDOI      \undefined \def \showDOI       #1{#1}\fi
\ifx \showISBNx    \undefined \def \showISBNx     #1{\unskip}     \fi
\ifx \showISBNxiii \undefined \def \showISBNxiii  #1{\unskip}     \fi
\ifx \showISSN     \undefined \def \showISSN      #1{\unskip}     \fi
\ifx \showLCCN     \undefined \def \showLCCN      #1{\unskip}     \fi
\ifx \shownote     \undefined \def \shownote      #1{#1}          \fi
\ifx \showarticletitle \undefined \def \showarticletitle #1{#1}   \fi
\ifx \showURL      \undefined \def \showURL       {\relax}        \fi
\providecommand\bibfield[2]{#2}
\providecommand\bibinfo[2]{#2}
\providecommand\natexlab[1]{#1}
\providecommand\showeprint[2][]{arXiv:#2}

\bibitem[Araslanov et~al\mbox{.}(2021)]%
        {Dense_araslanov2021dense}
\bibfield{author}{\bibinfo{person}{Nikita Araslanov}, \bibinfo{person}{Simone
  Schaub-Meyer}, {and} \bibinfo{person}{Stefan Roth}.}
  \bibinfo{year}{2021}\natexlab{}.
\newblock \showarticletitle{Dense Unsupervised Learning for Video
  Segmentation}.
\newblock \bibinfo{journal}{\emph{Advances in Neural Information Processing
  Systems}}  \bibinfo{volume}{34} (\bibinfo{year}{2021}).
\newblock


\bibitem[Bao et~al\mbox{.}(2021)]%
        {BEiT_bao2021beit}
\bibfield{author}{\bibinfo{person}{Hangbo Bao}, \bibinfo{person}{Li Dong},
  {and} \bibinfo{person}{Furu Wei}.} \bibinfo{year}{2021}\natexlab{}.
\newblock \showarticletitle{Beit: Bert pre-training of image transformers}.
\newblock \bibinfo{journal}{\emph{arXiv preprint arXiv:2106.08254}}
  (\bibinfo{year}{2021}).
\newblock


\bibitem[Carion et~al\mbox{.}(2020)]%
        {DERT_carion2020end}
\bibfield{author}{\bibinfo{person}{Nicolas Carion}, \bibinfo{person}{Francisco
  Massa}, \bibinfo{person}{Gabriel Synnaeve}, \bibinfo{person}{Nicolas
  Usunier}, \bibinfo{person}{Alexander Kirillov}, {and} \bibinfo{person}{Sergey
  Zagoruyko}.} \bibinfo{year}{2020}\natexlab{}.
\newblock \showarticletitle{End-to-end object detection with transformers}. In
  \bibinfo{booktitle}{\emph{European conference on computer vision}}. Springer,
  \bibinfo{pages}{213--229}.
\newblock


\bibitem[Caron et~al\mbox{.}(2021)]%
        {DINO_caron2021emerging}
\bibfield{author}{\bibinfo{person}{Mathilde Caron}, \bibinfo{person}{Hugo
  Touvron}, \bibinfo{person}{Ishan Misra}, \bibinfo{person}{Herv{\'e}
  J{\'e}gou}, \bibinfo{person}{Julien Mairal}, \bibinfo{person}{Piotr
  Bojanowski}, {and} \bibinfo{person}{Armand Joulin}.}
  \bibinfo{year}{2021}\natexlab{}.
\newblock \showarticletitle{Emerging properties in self-supervised vision
  transformers}. In \bibinfo{booktitle}{\emph{Proceedings of the IEEE/CVF
  International Conference on Computer Vision}}. \bibinfo{pages}{9650--9660}.
\newblock


\bibitem[Devlin et~al\mbox{.}(2018)]%
        {BERT_devlin2018bert}
\bibfield{author}{\bibinfo{person}{Jacob Devlin}, \bibinfo{person}{Ming-Wei
  Chang}, \bibinfo{person}{Kenton Lee}, {and} \bibinfo{person}{Kristina
  Toutanova}.} \bibinfo{year}{2018}\natexlab{}.
\newblock \showarticletitle{Bert: Pre-training of deep bidirectional
  transformers for language understanding}.
\newblock \bibinfo{journal}{\emph{arXiv preprint arXiv:1810.04805}}
  (\bibinfo{year}{2018}).
\newblock


\bibitem[Dosovitskiy et~al\mbox{.}(2020)]%
        {ViT_dosovitskiy2020image}
\bibfield{author}{\bibinfo{person}{Alexey Dosovitskiy}, \bibinfo{person}{Lucas
  Beyer}, \bibinfo{person}{Alexander Kolesnikov}, \bibinfo{person}{Dirk
  Weissenborn}, \bibinfo{person}{Xiaohua Zhai}, \bibinfo{person}{Thomas
  Unterthiner}, \bibinfo{person}{Mostafa Dehghani}, \bibinfo{person}{Matthias
  Minderer}, \bibinfo{person}{Georg Heigold}, \bibinfo{person}{Sylvain Gelly},
  {et~al\mbox{.}}} \bibinfo{year}{2020}\natexlab{}.
\newblock \showarticletitle{An image is worth 16x16 words: Transformers for
  image recognition at scale}.
\newblock \bibinfo{journal}{\emph{arXiv preprint arXiv:2010.11929}}
  (\bibinfo{year}{2020}).
\newblock


\bibitem[Goyal et~al\mbox{.}(2017)]%
        {LinearWarmUp_2017accurate}
\bibfield{author}{\bibinfo{person}{Priya Goyal}, \bibinfo{person}{Piotr
  Doll{\'a}r}, \bibinfo{person}{Ross Girshick}, \bibinfo{person}{Pieter
  Noordhuis}, \bibinfo{person}{Lukasz Wesolowski}, \bibinfo{person}{Aapo
  Kyrola}, \bibinfo{person}{Andrew Tulloch}, \bibinfo{person}{Yangqing Jia},
  {and} \bibinfo{person}{Kaiming He}.} \bibinfo{year}{2017}\natexlab{}.
\newblock \showarticletitle{Accurate, large minibatch sgd: Training imagenet in
  1 hour}.
\newblock \bibinfo{journal}{\emph{arXiv preprint arXiv:1706.02677}}
  (\bibinfo{year}{2017}).
\newblock


\bibitem[Grill et~al\mbox{.}(2020)]%
        {BYOL_grill2020bootstrap}
\bibfield{author}{\bibinfo{person}{Jean-Bastien Grill},
  \bibinfo{person}{Florian Strub}, \bibinfo{person}{Florent Altch{\'e}},
  \bibinfo{person}{Corentin Tallec}, \bibinfo{person}{Pierre Richemond},
  \bibinfo{person}{Elena Buchatskaya}, \bibinfo{person}{Carl Doersch},
  \bibinfo{person}{Bernardo Avila~Pires}, \bibinfo{person}{Zhaohan Guo},
  \bibinfo{person}{Mohammad Gheshlaghi~Azar}, {et~al\mbox{.}}}
  \bibinfo{year}{2020}\natexlab{}.
\newblock \showarticletitle{Bootstrap your own latent-a new approach to
  self-supervised learning}.
\newblock \bibinfo{journal}{\emph{Advances in Neural Information Processing
  Systems}}  \bibinfo{volume}{33} (\bibinfo{year}{2020}),
  \bibinfo{pages}{21271--21284}.
\newblock


\bibitem[He et~al\mbox{.}(2021)]%
        {MAE_he2021masked}
\bibfield{author}{\bibinfo{person}{Kaiming He}, \bibinfo{person}{Xinlei Chen},
  \bibinfo{person}{Saining Xie}, \bibinfo{person}{Yanghao Li},
  \bibinfo{person}{Piotr Doll{\'a}r}, {and} \bibinfo{person}{Ross Girshick}.}
  \bibinfo{year}{2021}\natexlab{}.
\newblock \showarticletitle{Masked autoencoders are scalable vision learners}.
\newblock \bibinfo{journal}{\emph{arXiv preprint arXiv:2111.06377}}
  (\bibinfo{year}{2021}).
\newblock


\bibitem[He et~al\mbox{.}(2020)]%
        {MOCO_he2020momentum}
\bibfield{author}{\bibinfo{person}{Kaiming He}, \bibinfo{person}{Haoqi Fan},
  \bibinfo{person}{Yuxin Wu}, \bibinfo{person}{Saining Xie}, {and}
  \bibinfo{person}{Ross Girshick}.} \bibinfo{year}{2020}\natexlab{}.
\newblock \showarticletitle{Momentum contrast for unsupervised visual
  representation learning}. In \bibinfo{booktitle}{\emph{Proceedings of the
  IEEE/CVF conference on computer vision and pattern recognition}}.
  \bibinfo{pages}{9729--9738}.
\newblock


\bibitem[He et~al\mbox{.}(2016)]%
        {ResNet_he2016deep}
\bibfield{author}{\bibinfo{person}{Kaiming He}, \bibinfo{person}{Xiangyu
  Zhang}, \bibinfo{person}{Shaoqing Ren}, {and} \bibinfo{person}{Jian Sun}.}
  \bibinfo{year}{2016}\natexlab{}.
\newblock \showarticletitle{Deep residual learning for image recognition}. In
  \bibinfo{booktitle}{\emph{Proceedings of the IEEE conference on computer
  vision and pattern recognition}}. \bibinfo{pages}{770--778}.
\newblock


\bibitem[Jabri et~al\mbox{.}(2020)]%
        {VRW_jabri2020space}
\bibfield{author}{\bibinfo{person}{Allan Jabri}, \bibinfo{person}{Andrew
  Owens}, {and} \bibinfo{person}{Alexei Efros}.}
  \bibinfo{year}{2020}\natexlab{}.
\newblock \showarticletitle{Space-time correspondence as a contrastive random
  walk}.
\newblock \bibinfo{journal}{\emph{Advances in neural information processing
  systems}}  \bibinfo{volume}{33} (\bibinfo{year}{2020}),
  \bibinfo{pages}{19545--19560}.
\newblock


\bibitem[Jeon et~al\mbox{.}(2021)]%
        {R3_added3_jeon2021mining}
\bibfield{author}{\bibinfo{person}{Sangryul Jeon}, \bibinfo{person}{Dongbo
  Min}, \bibinfo{person}{Seungryong Kim}, {and} \bibinfo{person}{Kwanghoon
  Sohn}.} \bibinfo{year}{2021}\natexlab{}.
\newblock \showarticletitle{Mining better samples for contrastive learning of
  temporal correspondence}. In \bibinfo{booktitle}{\emph{Proceedings of the
  IEEE/CVF Conference on Computer Vision and Pattern Recognition}}.
  \bibinfo{pages}{1034--1044}.
\newblock


\bibitem[Kay et~al\mbox{.}(2017)]%
        {Kinetics-400_2_kay2017kinetics}
\bibfield{author}{\bibinfo{person}{Will Kay}, \bibinfo{person}{Joao Carreira},
  \bibinfo{person}{Karen Simonyan}, \bibinfo{person}{Brian Zhang},
  \bibinfo{person}{Chloe Hillier}, \bibinfo{person}{Sudheendra
  Vijayanarasimhan}, \bibinfo{person}{Fabio Viola}, \bibinfo{person}{Tim
  Green}, \bibinfo{person}{Trevor Back}, \bibinfo{person}{Paul Natsev},
  {et~al\mbox{.}}} \bibinfo{year}{2017}\natexlab{}.
\newblock \showarticletitle{The kinetics human action video dataset}.
\newblock \bibinfo{journal}{\emph{arXiv preprint arXiv:1705.06950}}
  (\bibinfo{year}{2017}).
\newblock


\bibitem[Lai et~al\mbox{.}(2020)]%
        {MAST_lai2020mast}
\bibfield{author}{\bibinfo{person}{Zihang Lai}, \bibinfo{person}{Erika Lu},
  {and} \bibinfo{person}{Weidi Xie}.} \bibinfo{year}{2020}\natexlab{}.
\newblock \showarticletitle{MAST: A memory-augmented self-supervised tracker}.
  In \bibinfo{booktitle}{\emph{Proceedings of the IEEE/CVF Conference on
  Computer Vision and Pattern Recognition}}. \bibinfo{pages}{6479--6488}.
\newblock


\bibitem[Lai and Xie(2019)]%
        {CorrFlow_lai2019self}
\bibfield{author}{\bibinfo{person}{Zihang Lai} {and} \bibinfo{person}{Weidi
  Xie}.} \bibinfo{year}{2019}\natexlab{}.
\newblock \showarticletitle{Self-supervised learning for video correspondence
  flow}.
\newblock \bibinfo{journal}{\emph{arXiv preprint arXiv:1905.00875}}
  (\bibinfo{year}{2019}).
\newblock


\bibitem[Liang et~al\mbox{.}(2021)]%
        {VOS2_liang2021rethinking}
\bibfield{author}{\bibinfo{person}{Chen Liang}, \bibinfo{person}{Yu Wu},
  \bibinfo{person}{Tianfei Zhou}, \bibinfo{person}{Wenguan Wang},
  \bibinfo{person}{Zongxin Yang}, \bibinfo{person}{Yunchao Wei}, {and}
  \bibinfo{person}{Yi Yang}.} \bibinfo{year}{2021}\natexlab{}.
\newblock \showarticletitle{Rethinking cross-modal interaction from a top-down
  perspective for referring video object segmentation}.
\newblock \bibinfo{journal}{\emph{arXiv preprint arXiv:2106.01061}}
  (\bibinfo{year}{2021}).
\newblock


\bibitem[Lin et~al\mbox{.}(2021)]%
        {Video_editing_lin2021automated}
\bibfield{author}{\bibinfo{person}{Qin Lin}, \bibinfo{person}{Nuo Pang}, {and}
  \bibinfo{person}{Zhiying Hong}.} \bibinfo{year}{2021}\natexlab{}.
\newblock \showarticletitle{Automated Multi-Modal Video Editing for Ads Video}.
  In \bibinfo{booktitle}{\emph{Proceedings of the 29th ACM International
  Conference on Multimedia}}. \bibinfo{pages}{4823--4827}.
\newblock


\bibitem[Liu et~al\mbox{.}(2021)]%
        {SwinTrans_liu2021swin}
\bibfield{author}{\bibinfo{person}{Ze Liu}, \bibinfo{person}{Yutong Lin},
  \bibinfo{person}{Yue Cao}, \bibinfo{person}{Han Hu}, \bibinfo{person}{Yixuan
  Wei}, \bibinfo{person}{Zheng Zhang}, \bibinfo{person}{Stephen Lin}, {and}
  \bibinfo{person}{Baining Guo}.} \bibinfo{year}{2021}\natexlab{}.
\newblock \showarticletitle{Swin transformer: Hierarchical vision transformer
  using shifted windows}. In \bibinfo{booktitle}{\emph{Proceedings of the
  IEEE/CVF International Conference on Computer Vision}}.
  \bibinfo{pages}{10012--10022}.
\newblock


\bibitem[Loshchilov and Hutter(2016)]%
        {CosineScheduler_loshchilov2016sgdr}
\bibfield{author}{\bibinfo{person}{Ilya Loshchilov} {and}
  \bibinfo{person}{Frank Hutter}.} \bibinfo{year}{2016}\natexlab{}.
\newblock \showarticletitle{Sgdr: Stochastic gradient descent with warm
  restarts}.
\newblock \bibinfo{journal}{\emph{arXiv preprint arXiv:1608.03983}}
  (\bibinfo{year}{2016}).
\newblock


\bibitem[Loshchilov and Hutter(2018)]%
        {AdamW_loshchilov2018fixing}
\bibfield{author}{\bibinfo{person}{Ilya Loshchilov} {and}
  \bibinfo{person}{Frank Hutter}.} \bibinfo{year}{2018}\natexlab{}.
\newblock \showarticletitle{Fixing weight decay regularization in adam}.
\newblock  (\bibinfo{year}{2018}).
\newblock


\bibitem[Miao et~al\mbox{.}(2020)]%
        {VOS1_miao2020memory}
\bibfield{author}{\bibinfo{person}{Jiaxu Miao}, \bibinfo{person}{Yunchao Wei},
  {and} \bibinfo{person}{Yi Yang}.} \bibinfo{year}{2020}\natexlab{}.
\newblock \showarticletitle{Memory aggregation networks for efficient
  interactive video object segmentation}. In
  \bibinfo{booktitle}{\emph{Proceedings of the IEEE/CVF Conference on Computer
  Vision and Pattern Recognition}}. \bibinfo{pages}{10366--10375}.
\newblock


\bibitem[Muller et~al\mbox{.}(2018)]%
        {TrackingNet_Muller_2018_ECCV}
\bibfield{author}{\bibinfo{person}{Matthias Muller}, \bibinfo{person}{Adel
  Bibi}, \bibinfo{person}{Silvio Giancola}, \bibinfo{person}{Salman Alsubaihi},
  {and} \bibinfo{person}{Bernard Ghanem}.} \bibinfo{year}{2018}\natexlab{}.
\newblock \showarticletitle{TrackingNet: A Large-Scale Dataset and Benchmark
  for Object Tracking in the Wild}. In \bibinfo{booktitle}{\emph{The European
  Conference on Computer Vision (ECCV)}}.
\newblock


\bibitem[Oh et~al\mbox{.}(2019)]%
        {STM_oh2019video}
\bibfield{author}{\bibinfo{person}{Seoung~Wug Oh}, \bibinfo{person}{Joon-Young
  Lee}, \bibinfo{person}{Ning Xu}, {and} \bibinfo{person}{Seon~Joo Kim}.}
  \bibinfo{year}{2019}\natexlab{}.
\newblock \showarticletitle{Video object segmentation using space-time memory
  networks}. In \bibinfo{booktitle}{\emph{Proceedings of the IEEE/CVF
  International Conference on Computer Vision}}. \bibinfo{pages}{9226--9235}.
\newblock


\bibitem[Perazzi et~al\mbox{.}(2016)]%
        {JF_perazzi2016benchmark}
\bibfield{author}{\bibinfo{person}{Federico Perazzi}, \bibinfo{person}{Jordi
  Pont-Tuset}, \bibinfo{person}{Brian McWilliams}, \bibinfo{person}{Luc
  Van~Gool}, \bibinfo{person}{Markus Gross}, {and} \bibinfo{person}{Alexander
  Sorkine-Hornung}.} \bibinfo{year}{2016}\natexlab{}.
\newblock \showarticletitle{A benchmark dataset and evaluation methodology for
  video object segmentation}. In \bibinfo{booktitle}{\emph{Proceedings of the
  IEEE conference on computer vision and pattern recognition}}.
  \bibinfo{pages}{724--732}.
\newblock


\bibitem[Pont-Tuset et~al\mbox{.}(2017)]%
        {DAVIS2017_pont20172017}
\bibfield{author}{\bibinfo{person}{Jordi Pont-Tuset}, \bibinfo{person}{Federico
  Perazzi}, \bibinfo{person}{Sergi Caelles}, \bibinfo{person}{Pablo
  Arbel{\'a}ez}, \bibinfo{person}{Alex Sorkine-Hornung}, {and}
  \bibinfo{person}{Luc Van~Gool}.} \bibinfo{year}{2017}\natexlab{}.
\newblock \showarticletitle{The 2017 davis challenge on video object
  segmentation}.
\newblock \bibinfo{journal}{\emph{arXiv preprint arXiv:1704.00675}}
  (\bibinfo{year}{2017}).
\newblock


\bibitem[Sigurdsson et~al\mbox{.}(2016)]%
        {Charades_sigurdsson2016hollywood}
\bibfield{author}{\bibinfo{person}{Gunnar~A Sigurdsson},
  \bibinfo{person}{G{\"u}l Varol}, \bibinfo{person}{Xiaolong Wang},
  \bibinfo{person}{Ali Farhadi}, \bibinfo{person}{Ivan Laptev}, {and}
  \bibinfo{person}{Abhinav Gupta}.} \bibinfo{year}{2016}\natexlab{}.
\newblock \showarticletitle{Hollywood in homes: Crowdsourcing data collection
  for activity understanding}. In \bibinfo{booktitle}{\emph{European Conference
  on Computer Vision}}. Springer, \bibinfo{pages}{510--526}.
\newblock


\bibitem[Touvron et~al\mbox{.}(2021)]%
        {DeiT_touvron2021training}
\bibfield{author}{\bibinfo{person}{Hugo Touvron}, \bibinfo{person}{Matthieu
  Cord}, \bibinfo{person}{Matthijs Douze}, \bibinfo{person}{Francisco Massa},
  \bibinfo{person}{Alexandre Sablayrolles}, {and} \bibinfo{person}{Herv{\'e}
  J{\'e}gou}.} \bibinfo{year}{2021}\natexlab{}.
\newblock \showarticletitle{Training data-efficient image transformers \&
  distillation through attention}. In \bibinfo{booktitle}{\emph{International
  Conference on Machine Learning}}. PMLR, \bibinfo{pages}{10347--10357}.
\newblock


\bibitem[Valmadre et~al\mbox{.}(2018)]%
        {OxUvA_valmadre2018long}
\bibfield{author}{\bibinfo{person}{Jack Valmadre}, \bibinfo{person}{Luca
  Bertinetto}, \bibinfo{person}{Joao~F Henriques}, \bibinfo{person}{Ran Tao},
  \bibinfo{person}{Andrea Vedaldi}, \bibinfo{person}{Arnold~WM Smeulders},
  \bibinfo{person}{Philip~HS Torr}, {and} \bibinfo{person}{Efstratios Gavves}.}
  \bibinfo{year}{2018}\natexlab{}.
\newblock \showarticletitle{Long-term tracking in the wild: A benchmark}. In
  \bibinfo{booktitle}{\emph{Proceedings of the European conference on computer
  vision (ECCV)}}. \bibinfo{pages}{670--685}.
\newblock


\bibitem[Vondrick et~al\mbox{.}(2018)]%
        {Colorize_vondrick2018tracking}
\bibfield{author}{\bibinfo{person}{Carl Vondrick}, \bibinfo{person}{Abhinav
  Shrivastava}, \bibinfo{person}{Alireza Fathi}, \bibinfo{person}{Sergio
  Guadarrama}, {and} \bibinfo{person}{Kevin Murphy}.}
  \bibinfo{year}{2018}\natexlab{}.
\newblock \showarticletitle{Tracking emerges by colorizing videos}. In
  \bibinfo{booktitle}{\emph{Proceedings of the European conference on computer
  vision (ECCV)}}. \bibinfo{pages}{391--408}.
\newblock


\bibitem[Wang et~al\mbox{.}(2020)]%
        {ContrastCorr_wang2020contrastive}
\bibfield{author}{\bibinfo{person}{Ning Wang}, \bibinfo{person}{Wengang Zhou},
  {and} \bibinfo{person}{Houqiang Li}.} \bibinfo{year}{2020}\natexlab{}.
\newblock \showarticletitle{Contrastive transformation for self-supervised
  correspondence learning}.
\newblock \bibinfo{journal}{\emph{arXiv preprint arXiv:2012.05057}}
  (\bibinfo{year}{2020}).
\newblock


\bibitem[Wang et~al\mbox{.}(2021b)]%
        {R4_added5_wang2021survey}
\bibfield{author}{\bibinfo{person}{Wenguan Wang}, \bibinfo{person}{Tianfei
  Zhou}, \bibinfo{person}{Fatih Porikli}, \bibinfo{person}{David Crandall},
  {and} \bibinfo{person}{Luc Van~Gool}.} \bibinfo{year}{2021}\natexlab{b}.
\newblock \showarticletitle{A survey on deep learning technique for video
  segmentation}.
\newblock \bibinfo{journal}{\emph{arXiv preprint arXiv:2107.01153}}
  (\bibinfo{year}{2021}).
\newblock


\bibitem[Wang et~al\mbox{.}(2019)]%
        {TimeCycle_wang2019learning}
\bibfield{author}{\bibinfo{person}{Xiaolong Wang}, \bibinfo{person}{Allan
  Jabri}, {and} \bibinfo{person}{Alexei~A Efros}.}
  \bibinfo{year}{2019}\natexlab{}.
\newblock \showarticletitle{Learning correspondence from the cycle-consistency
  of time}. In \bibinfo{booktitle}{\emph{Proceedings of the IEEE/CVF Conference
  on Computer Vision and Pattern Recognition}}. \bibinfo{pages}{2566--2576}.
\newblock


\bibitem[Wang et~al\mbox{.}(2021a)]%
        {VisTR_wang2021end}
\bibfield{author}{\bibinfo{person}{Yuqing Wang}, \bibinfo{person}{Zhaoliang
  Xu}, \bibinfo{person}{Xinlong Wang}, \bibinfo{person}{Chunhua Shen},
  \bibinfo{person}{Baoshan Cheng}, \bibinfo{person}{Hao Shen}, {and}
  \bibinfo{person}{Huaxia Xia}.} \bibinfo{year}{2021}\natexlab{a}.
\newblock \showarticletitle{End-to-end video instance segmentation with
  transformers}. In \bibinfo{booktitle}{\emph{Proceedings of the IEEE/CVF
  Conference on Computer Vision and Pattern Recognition}}.
  \bibinfo{pages}{8741--8750}.
\newblock


\bibitem[Xiong et~al\mbox{.}(2021)]%
        {VR_xiong2021augmented}
\bibfield{author}{\bibinfo{person}{Jianghao Xiong}, \bibinfo{person}{En-Lin
  Hsiang}, \bibinfo{person}{Ziqian He}, \bibinfo{person}{Tao Zhan}, {and}
  \bibinfo{person}{Shin-Tson Wu}.} \bibinfo{year}{2021}\natexlab{}.
\newblock \showarticletitle{Augmented reality and virtual reality displays:
  emerging technologies and future perspectives}.
\newblock \bibinfo{journal}{\emph{Light: Science \& Applications}}
  \bibinfo{volume}{10}, \bibinfo{number}{1} (\bibinfo{year}{2021}),
  \bibinfo{pages}{1--30}.
\newblock


\bibitem[Xu and Wang(2021)]%
        {Rethinking_xu2021rethinking}
\bibfield{author}{\bibinfo{person}{Jiarui Xu} {and} \bibinfo{person}{Xiaolong
  Wang}.} \bibinfo{year}{2021}\natexlab{}.
\newblock \showarticletitle{Rethinking self-supervised correspondence learning:
  A video frame-level similarity perspective}. In
  \bibinfo{booktitle}{\emph{Proceedings of the IEEE/CVF International
  Conference on Computer Vision}}. \bibinfo{pages}{10075--10085}.
\newblock


\bibitem[Xu et~al\mbox{.}(2018)]%
        {YouTube-Vos_xu2018youtube}
\bibfield{author}{\bibinfo{person}{Ning Xu}, \bibinfo{person}{Linjie Yang},
  \bibinfo{person}{Yuchen Fan}, \bibinfo{person}{Jianchao Yang},
  \bibinfo{person}{Dingcheng Yue}, \bibinfo{person}{Yuchen Liang},
  \bibinfo{person}{Brian Price}, \bibinfo{person}{Scott Cohen}, {and}
  \bibinfo{person}{Thomas Huang}.} \bibinfo{year}{2018}\natexlab{}.
\newblock \showarticletitle{Youtube-vos: Sequence-to-sequence video object
  segmentation}. In \bibinfo{booktitle}{\emph{Proceedings of the European
  conference on computer vision (ECCV)}}. \bibinfo{pages}{585--601}.
\newblock


\bibitem[Yang et~al\mbox{.}(2019b)]%
        {R2_added0_yang2019unsupervised}
\bibfield{author}{\bibinfo{person}{Yanchao Yang}, \bibinfo{person}{Antonio
  Loquercio}, \bibinfo{person}{Davide Scaramuzza}, {and}
  \bibinfo{person}{Stefano Soatto}.} \bibinfo{year}{2019}\natexlab{b}.
\newblock \showarticletitle{Unsupervised moving object detection via contextual
  information separation}. In \bibinfo{booktitle}{\emph{Proceedings of the
  IEEE/CVF Conference on Computer Vision and Pattern Recognition}}.
  \bibinfo{pages}{879--888}.
\newblock


\bibitem[Yang et~al\mbox{.}(2019a)]%
        {VOS3_yang2019going}
\bibfield{author}{\bibinfo{person}{Zongxin Yang}, \bibinfo{person}{Peike Li},
  \bibinfo{person}{Qianyu Feng}, \bibinfo{person}{Yunchao Wei}, {and}
  \bibinfo{person}{Yi Yang}.} \bibinfo{year}{2019}\natexlab{a}.
\newblock \showarticletitle{Going deeper into embedding learning for video
  object segmentation}. In \bibinfo{booktitle}{\emph{Proceedings of the
  IEEE/CVF International Conference on Computer Vision Workshops}}.
  \bibinfo{pages}{0--0}.
\newblock


\bibitem[Yang et~al\mbox{.}(2020)]%
        {CFBI_eccv_yang2020collaborative}
\bibfield{author}{\bibinfo{person}{Zongxin Yang}, \bibinfo{person}{Yunchao
  Wei}, {and} \bibinfo{person}{Yi Yang}.} \bibinfo{year}{2020}\natexlab{}.
\newblock \showarticletitle{Collaborative video object segmentation by
  foreground-background integration}. In \bibinfo{booktitle}{\emph{European
  Conference on Computer Vision}}. Springer, \bibinfo{pages}{332--348}.
\newblock


\bibitem[Yang et~al\mbox{.}(2021a)]%
        {yang2021associating}
\bibfield{author}{\bibinfo{person}{Zongxin Yang}, \bibinfo{person}{Yunchao
  Wei}, {and} \bibinfo{person}{Yi Yang}.} \bibinfo{year}{2021}\natexlab{a}.
\newblock \showarticletitle{Associating objects with transformers for video
  object segmentation}.
\newblock \bibinfo{journal}{\emph{Advances in Neural Information Processing
  Systems}}  \bibinfo{volume}{34} (\bibinfo{year}{2021}),
  \bibinfo{pages}{2491--2502}.
\newblock


\bibitem[Yang et~al\mbox{.}(2021b)]%
        {CFBI_yang2021collaborative}
\bibfield{author}{\bibinfo{person}{Zongxin Yang}, \bibinfo{person}{Yunchao
  Wei}, {and} \bibinfo{person}{Yi Yang}.} \bibinfo{year}{2021}\natexlab{b}.
\newblock \showarticletitle{Collaborative video object segmentation by
  multi-scale foreground-background integration}.
\newblock \bibinfo{journal}{\emph{IEEE Transactions on Pattern Analysis and
  Machine Intelligence}} (\bibinfo{year}{2021}).
\newblock


\bibitem[Zhao et~al\mbox{.}(2021)]%
        {R3_added4_zhao2021modelling}
\bibfield{author}{\bibinfo{person}{Zixu Zhao}, \bibinfo{person}{Yueming Jin},
  {and} \bibinfo{person}{Pheng-Ann Heng}.} \bibinfo{year}{2021}\natexlab{}.
\newblock \showarticletitle{Modelling neighbor relation in joint space-time
  graph for video correspondence learning}. In
  \bibinfo{booktitle}{\emph{Proceedings of the IEEE/CVF International
  Conference on Computer Vision}}. \bibinfo{pages}{9960--9969}.
\newblock


\bibitem[Zhou et~al\mbox{.}(2021)]%
        {iBOT_zhou2021ibot}
\bibfield{author}{\bibinfo{person}{Jinghao Zhou}, \bibinfo{person}{Chen Wei},
  \bibinfo{person}{Huiyu Wang}, \bibinfo{person}{Wei Shen},
  \bibinfo{person}{Cihang Xie}, \bibinfo{person}{Alan Yuille}, {and}
  \bibinfo{person}{Tao Kong}.} \bibinfo{year}{2021}\natexlab{}.
\newblock \showarticletitle{ibot: Image bert pre-training with online
  tokenizer}.
\newblock \bibinfo{journal}{\emph{arXiv preprint arXiv:2111.07832}}
  (\bibinfo{year}{2021}).
\newblock


\end{thebibliography}
